\documentclass{article}

    \usepackage[preprint]{neurips_2022}

\usepackage[utf8]{inputenc} % allow utf-8 input
\usepackage[T1]{fontenc}    % use 8-bit T1 fonts
\usepackage[pagebackref,breaklinks,colorlinks]{hyperref}      % hyperlinks
\usepackage{url}            % simple URL typesetting
\usepackage{booktabs}       % professional-quality tables
\usepackage{amsfonts}       % blackboard math symbols
\usepackage{nicefrac}       % compact symbols for 1/2, etc.
\usepackage{microtype}      % microtypography
\usepackage{xcolor}         % colors

\usepackage{amsmath}
\usepackage{bm}
\usepackage{bbm}
\usepackage{graphicx}
\usepackage{subcaption}

\usepackage{caption}
\usepackage{graphicx, amsmath, amssymb, caption, subcaption, multirow, overpic, textpos}
\usepackage{enumitem}
\setlist[itemize]{topsep={0pt},partopsep={0pt}}

\usepackage{xspace}
\providecommand{\eg}{\textit{e.g.}\@\xspace}
\providecommand{\ie}{\textit{i.e.}\@\xspace}

\newlength\savewidth
\newcommand{\tablestyle}[2]{\setlength{\tabcolsep}{#1}\renewcommand{\arraystretch}{#2}\centering\footnotesize}

\title{Pretraining is All You Need for Image-to-Image Translation}

\author{Tengfei Wang$^1$\footnotemark[1] \qquad
Ting Zhang$^2$ \qquad
Bo Zhang$^2$  \qquad
Hao Ouyang$^1$ 
\\
\textbf{Dong Chen}$^2$ \qquad
\textbf{Qifeng Chen}$^1$  \qquad
\textbf{Fang Wen}$^2$  
\\
\\
$^1$The Hong Kong University of Science and Technology \qquad
$^2$Microsoft Research Asia 
}

\begin{document}

\maketitle
\footnotetext[1]{Author did this work  during his internship at Microsoft Research Asia.}

\begin{abstract}
We propose to use pretraining to boost general image-to-image translation. Prior image-to-image translation methods usually need dedicated architectural design and train individual translation models from scratch, struggling for high-quality generation of complex scenes, especially when paired training data are not abundant. In this paper, we regard each image-to-image translation problem as a downstream task and introduce a simple and generic framework that adapts a pretrained diffusion model to accommodate various kinds of image-to-image translation. We also propose adversarial training to enhance the texture synthesis in the diffusion model training, in conjunction with normalized guidance sampling to improve the generation quality. We present extensive empirical comparison across various tasks on challenging benchmarks such as ADE20K, COCO-Stuff, and DIODE, showing the proposed pretraining-based image-to-image translation (PITI) is capable of synthesizing images of unprecedented realism and faithfulness. Code will be available on the \href{https://tengfei-wang.github.io/PITI/index.html}{project webpage}.
\end{abstract}

\section{Introduction}

\begin{figure*}[t]
  \centering
  \small
  \setlength\tabcolsep{1pt}
  \renewcommand{\arraystretch}{0.6}
    \includegraphics[width=\linewidth]{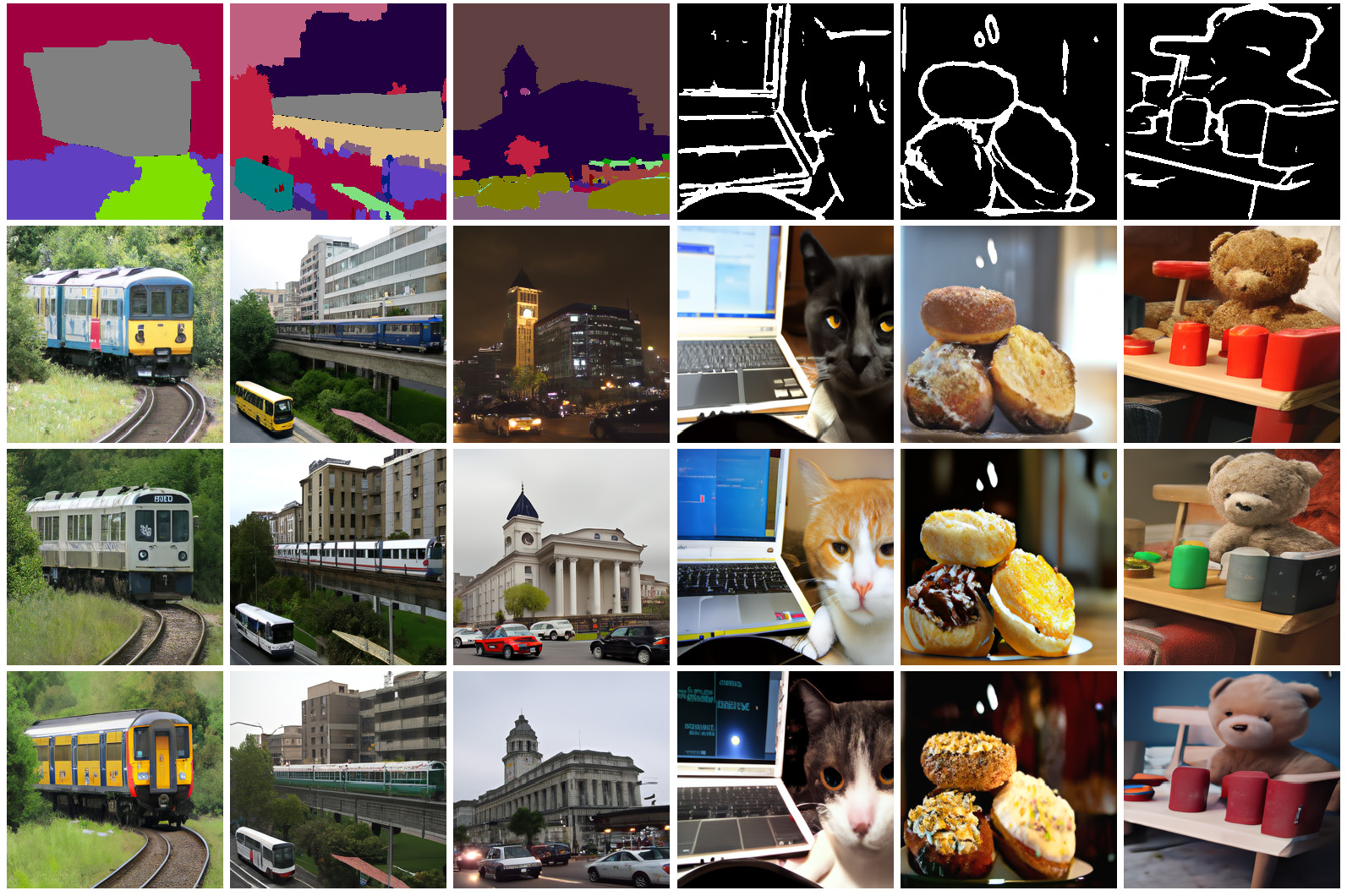}
  \caption{Diverse images sampled by our method given semantic layouts or sketches.}
  \label{fig:teaser}
\end{figure*}

Many content creation tasks involve converting an input image, \eg, a casual drawing, to a photo-realistic output. Such an image-to-image translation problem~\cite{isola2017image} essentially relates to learning the conditional distribution of natural images given the input using deep generative models. Over the years, we have seen a plethora of methods~\cite{pang2021image,liu2020generative} with task-specific customization that steadily pushes the state of the arts, yet it remains challenging for existing solutions to produce high-fidelity images satisfying practical usage.

Motivated by the tremendous success of network pretraining in various vision tasks~\cite{he2020momentum,chen2020simple,he2021masked,radford2021learning} and natural language processing~\cite{devlin2018bert,brown2020language}, we propose a new paradigm that uses pretraining to improve image-to-image translation. The key idea is to use a pretrained neural network to capture the natural image manifold, and thus the image translation is equivalent to traversing this manifold and finding the feasible point that relates to the input semantics. Specifically, the synthesis network should be pretrained using a massive amount of images and serves as a generative prior that any sampling from its latent space will lead to a plausible output. With a capable pretrained synthesis network, the downstream training simply adapts the user input to the latent representation recognizable by the pretrained model. Compared to prior works that compromise the image quality to suit the prescribed semantic layout, the proposed framework guarantees the translation quality since the produced samples will rigorously lie on the natural image manifold.

The generative prior should possess the following properties. First, the pretrained model should have a strong capability to model complex scenes and ideally capture the whole natural image distribution. Rather than using GANs~\cite{goodfellow2014generative,brock2018large,karras2019style,zhang2021styleswin} that mainly work for specific domains (\eg, faces), we opt to use the diffusion model~\cite{ho2020denoising,song2020improved,dhariwal2021diffusion} which emerges to show impressive expressivity of synthesizing a wide variety of images. Second, it is expected to generate images from two kinds of latent codes: one characterizes the image semantics while the other accounts for the remaining image variations. In particular, a semantic and low-dimensional latent is pivotal for downstream tasks, otherwise it will be difficult to map distinct modality inputs to a complicated latent space. In view of these, we adopt GLIDE~\cite{nichol2021glide} as our pretrained generative prior, which is a diffusion model trained on huge data and can faithfully generate diverse images. Since the GLIDE model uses the latent corresponding to the text condition, it naturally admits a desired semantic latent space.

In order to accommodate the downstream tasks, we train a task-specific head that projects the translation input, \eg, segmentation mask, to the latent space of the pretrained model. Hence, the network for downstream tasks adopts an encoder-decoder architecture: the encoder translates the input to a task-agnostic latent space, followed by a powerful decoder, \ie, diffusion model, to produce a plausible image accordingly. In practice, we first fix the pretrained decoder and only update the encoder, and then we finetune the whole network jointly. Such stage-wise training can maximally utilize the pretrained knowledge while ensuring faithfulness to the given input.

We further propose techniques to improve the generation quality for the diffusion model. 1) We adopt the hierarchical generation strategy~\cite{ho2022cascaded,nichol2021glide,ramesh2022hierarchical}, which generates a coarse image and then performs super-resolution. However, we observe that a diffusion upsampler tends to produce oversmoothed results due to the Gaussian noise assumption in the denoising diffusion step and therefore introduce adversarial training during the denoising process, considerably enhancing the perceptual quality. 2) The commonly-used classifier-free guidance~\cite{ho2021classifier} leads to excessively saturated images with details washed away. To solve this, we propose to normalize the noise statistics explicitly. Such normalized guidance sampling allows more aggressive guidance and yields boosted generation quality.

Our {{p}retraining-based {i}mage-{t}o-{i}mage translation}, referred as \emph{PITI}, achieves unprecedented quality across a variety of downstream tasks, \eg, mask-to-image, sketch-to-image and geometry-to-image translation. Figure~\ref{fig:teaser} showcases some generated image samples of complex scenes which exhibit compelling quality and large diversity. Extensive experiments on challenging datasets, including ADE20K~\cite{zhou2017scene}, COCO-Stuff~\cite{caesar2018coco} and DIODE~\cite{vasiljevic2019diode}, show the significant superiority of our approach, as measured by both
quantitative metrics and subjective evaluation, over the state of the arts as well as the model without pretraining. Moreover, the proposed method shows promising potential for few-shot image-to-image translation.

\section{Related Work}
\paragraph{Image-to-image translation.} The goal is to synthesize images in the target domain while faithfully following the semantics of input. Plenty of works~\cite{pang2021image,liu2020generative} have been proposed to tackle the problem. The most popular choice is to use conditional generative adversarial networks~\cite{isola2017image,zhu2017toward,wang2018high,park2019semantic,choi2018stargan,zhang2020cross,zhou2021cocosnet} (cGAN) that rely on a discriminator to examine the gap with real images. More recently, autoregressive models~\cite{ramesh2021zero,esser2021taming} have shown promising results thanks to the outstanding expressivity of transformers, but they are slow to inference and prone to overfit. While remarkable progress has been achieved, these methods treat distinct tasks separately and have to learn from scratch using limited task-specific training data. Considering the synergy among tasks, some research efforts ~\cite{xia2021tedigan,zhang2021m6,kutuzova2021multimodal,huang2021multimodal,qian2019trinity,chen2021pre,saharia2021palette} aim to learn a unified model for diverse translation tasks via multi-task training. Differently in this paper, we propose to leverage the pretrained generative prior of general images and regard all the specific problems as downstream tasks. Unprecedented quality can be achieved since all the tasks can benefit from the pretrained knowledge about the natural images.

\paragraph{Image pretraining.}
It has proven crucial for contemporary vision tasks~\cite{he2021masked,bao2021beit,wei2021masked,baevski2022data2vec,radford2021learning} to first pretrain models on large data and then transfer the learned knowledge to downstream tasks. Nonetheless, a large research focus has been on discriminate tasks, while the pretraining for visual synthesis is much less explored.
There are prior attempts~\cite{patashnik2021styleclip,collins2020editing,yang2021gan, wang2021HFGI,wang2021towards,pan2021exploiting,menon2020pulse} to leverage a pretrained model as generative prior for conditional image synthesis, image editing and restoration. A GAN latent space~\cite{brock2018large,karras2019style} is often utilized, which embeds the input semantics and allows meaningful manipulation. However, GANs are only good at modeling specific image classes and suffer from mode dropping and stability issues. Hence, they are insufficient to serve as generative prior for general images. Recently SDEdit~\cite{meng2021sdedit} takes advantage of a powerful diffusion model but only targets stroke-to-image translation. In comparison, what we propose is a generic framework that uses pretraining to benefit various translation tasks, without any task-specific customization or hyper-parameter tuning.

\paragraph{Diffusion models.}
Recently diffusion and score-based models~\cite{sohl2015deep,ho2020denoising,song2020improved,dhariwal2021diffusion,ho2022cascaded} emerge to show competitive generation quality across various benchmarks. Notably, on the class-conditional ImageNet generation these models have already rivaled GAN-based methods in terms of both visual quality and sampling diversity. More recently, diffusion models have demonstrated extraordinary capacity when trained with large-scale text-image pairs~\cite{nichol2021glide,gu2021vector,ramesh2022hierarchical}. Saharia et al.~\cite{saharia2021palette} demonstrate the potential of using the diffusion model for image-to-image translation, but they only show results for data-rich problems, \eg, image colorization. Our work builds on these key advances, and we show how a well-pretrained diffusion model can serve as a universal generative prior that facilitates various synthesis tasks. On top of this, we propose instrumental techniques to improve the diffusion model on detailed texture synthesis as well as sampling quality.

\section{Approach}
\begin{figure}[t]
  \centering
  \includegraphics[width=\linewidth]{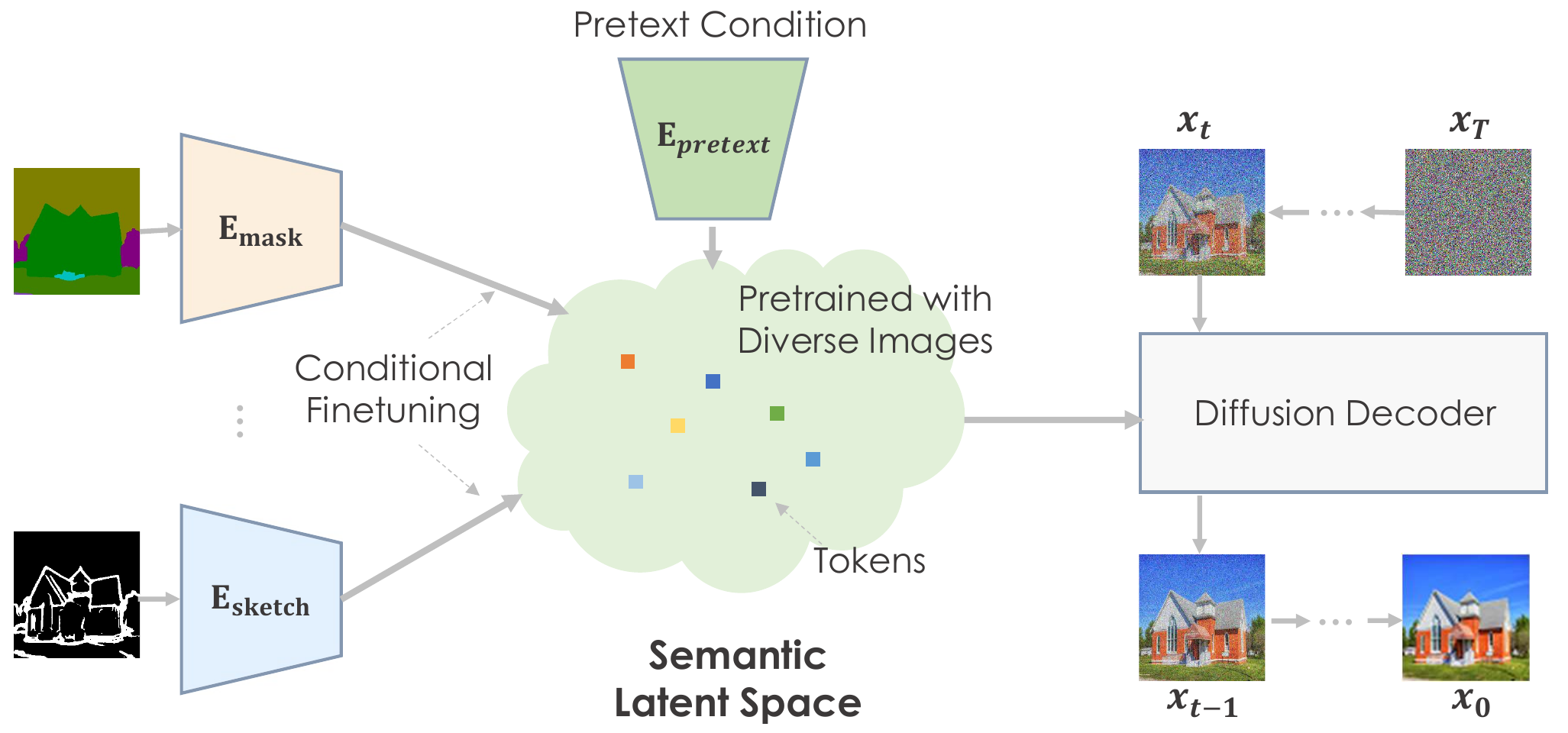}
  \caption{The overall framework. We can perform pretraining on huge data via different pretext tasks and learn a highly semantic latent space that models general and high-quality image statistics. For downstream tasks, we perform conditional finetuning to  map the task-specific conditions to this pretrained semantic space. By leveraging the pretrained knowledge, our model renders plausible images based on different conditions.}
%   \vspace{-1mm}
  \label{fig:overview}
\end{figure}

\subsection{Preliminary}
Diffusion models~\cite{sohl2015deep,ho2020denoising} iteratively produce an image by reversing a gradual noising process. The forward process $q$ corrupts the image $\bm{x}_0 \sim q(\bm{x}_0)$ by  gradually adding Gaussian noises in $T$ steps:
\begin{equation}
  q\left(\bm{x}_{t} | \bm{x}_{t-1}\right)= \mathcal{N}\left(\bm{x}_{t} ; \sqrt{1-\beta_{t}} \bm{x}_{t-1}, \beta_{t} \bm{I}\right),
\end{equation}
where $\beta_t$ determines the variance of noises added at each iteration. Thus, the forward process yields a sequence of increasingly noisy latent variables $\bm{x}_1,..., \bm{x}_T$, and after sufficient noising steps we reach a pure noise, \ie, $\bm{x}_T \sim \mathcal{N}(\bm{0},\bm{I})$. Importantly, one can marginalize out the intermediate steps and derive $\bm{x}_t$ from $\bm{x}_0$ directly, \ie,
\begin{equation}
    q\left(\bm{x}_{t} | \bm{x}_{0}\right) = \mathcal{N}\left(\bm{x}_{t} ; \sqrt{{\alpha}_{t}} \bm{x}_{0},\left(1-{\alpha}_{t}\right) \bm{I}\right),
\end{equation}
where $\alpha_{t}:=\prod_{i=1}^{t}\left(1-\beta_{i}\right)$. Or equivalently, we have $\bm{x}_{t}=\sqrt{{\alpha}_{t}} \bm{x}_{0}+\sqrt{1-{\alpha}_{t}} \bm{\epsilon}$, where $\bm{\epsilon}$ is the standard Gaussian noise.

To generate images from the data distribution, we can train a denoising model that starts from the Gaussian noise $\bm{x}_T \sim \mathcal{N}(\bm{0},\bm{I})$ and iteratively reduces the noise in the sequence of $\bm{x}_{T-1}, ..., \bm{x}_1, \bm{x}_0$. The denoising model $\bm{\epsilon}_{\theta}(\bm{x}_{t},t)$ takes the noisy input $\bm{x}_t$ at the timestep $t$ and predicts the added noise~$\bm{\epsilon}$ using a mean square error loss:
\begin{equation}
  \mathcal{L}_{\textnormal {simple }}=\mathbb{E}_{t, \bm{x}_0, \bm{\epsilon} \sim \mathcal{N}(0,I)} \Big[\big\|\bm{\epsilon}_{\theta}(\underbrace{\sqrt{\alpha_{t}} \bm{x}_0+\sqrt{1-\alpha_{t}}\bm{\epsilon}}_{\bm{x}_t} ,  t) - \bm{\epsilon}\big\|_2^2 \Big] .
\end{equation}

This reduction of image generation to denoising can be justified as the denoising score matching~\cite{song2020denoising} since $\nabla_{\bm{x}_{t}} \log p\left(\bm{x}_{t}\right) \propto \bm{\epsilon}_{\theta}(\bm{x}_t) $, or optimizing a simplified variational
lower bound of the data log-likelihood~\cite{ho2020denoising}. 

To ease the reverse diffusion process, one can additionally provide the condition $\bm{y}$, \eg, a class label, the text prompt or a degraded image. The denoising model hence becomes $\bm{\epsilon}_{\theta}(\bm{x}_t, \bm{y}, t)$ where the condition is injected through input concatenation~\cite{saharia2021image}, denormalization~\cite{dhariwal2021diffusion} or cross-attention~\cite{nichol2021glide}. 

Due to the striking generation capability on a wide range of images, diffusion models become an ideal choice to serve as generative prior. In the next, we will show how to properly pretrain the network using large data and then apply the learned knowledge to downstream tasks, as illustrated in Figure~\ref{fig:overview}.

\subsection{Generative pretraining}

As opposed to taking images from the same domain as for discriminate tasks, the pretrained model for generative tasks consumes vastly different kinds of images in distinct downstream tasks. Hence, during the generative pretraining, we expect the diffusion model to generate images from a latent space that is later shared to use for all the downstream tasks. Importantly, the pretrained model is desired to have a highly semantic space, \ie, neighboring points in this space corresponding to semantically similar images. In this way, the downstream finetuning only involves understanding the task-specific inputs while the challenging image synthesis --- rendering a plausible layout and realistic textures --- is accomplished using the pretrained knowledge.

To achieve this, we propose to pretrain the diffusion model to condition on a semantic input. Inspired by the impressive transferable capability of visual-linguistic pretraining~\cite{radford2021learning}, we adopt the GLIDE model~\cite{nichol2021glide} which is text-conditioned and is trained on huge and diverse text-image pairs. Specifically, a transformer network encodes the text input and produces text tokens that are further injected into the diffusion model. The textual embedding space is inherently semantic as desired. Similar to many recent works~\cite{ho2022cascaded,ramesh2022hierarchical}, GLIDE leverages a hierarchical generation scheme which begins with a \emph{base diffusion model} at the resolution of $64\times64$, followed by a \emph{diffusion upsampling model} to go from $64\times64$ to $256\times256$ resolution. Our experiment builds on the public GLIDE model, which is trained on approximately 67M text-image pairs with people and violent objects removed.

\subsection{Downstream adaptation}
Once the model is pretrained, we can adapt it to various downstream image synthesis tasks by using different strategies to finetune the base model and the upsampler model, respectively.

\textbf{Base model finetuning.} 
The generation using the base model can be formulated as $\bm{x}_t = \tilde{\mathcal{D}}\big(\tilde{\mathcal{E}}(\bm{x}_0, {\bm{y}})\big)$, where $\tilde{\mathcal{E}}$ and $\tilde{\mathcal{D}}$ denote the pretrained encoder and decoder respectively and ${\bm{y}}$ is the condition used for pretraining. In order to accommodate new modality conditions beyond texts, we train a task-specific head $\mathcal{E}_i$ to map the conditional input into the pretrained embedding space. If the input can be faithfully projected, the pretrained decoder will produce a plausible output.   

We propose a two-stage finetuning scheme. In the first stage, we specifically train the task-specific encoder and leave the pretrained decoder intact. The outputs at this stage will roughly match the semantics of the input, but without accurate spatial alignment. Then we finetune both the encoder and decoder altogether. After this, we obtain much improved spatial semantic alignment. Such stage-wise training is helpful to cultivate the pretrained knowledge as much as possible and is proven crucial for much improved quality.

\textbf{Adversarial diffusion upsampler.}
We further finetune the diffusion upsampler for high-resolution generation. Following~\cite{ho2022cascaded,ramesh2022hierarchical}, we apply random degradation, specifically the real-world BSR degradation~\cite{zhang2021designing}, on the training inputs to reduce the gap between training images and the samples from the base model. 
In particular, we also introduce $L_0$ filter~\cite{xu2011image} to mimic the oversmoothed effect.

Nonetheless, we still observe oversmoothed results even though we apply strong data augmentations. We conjecture the issue arises from the Gaussian noise assumption in the diffusion denoising processing. Hence, besides computing a standard mean square error loss for noise prediction, we propose to impose perceptual loss~\cite{johnson2016perceptual} and adversarial loss~\cite{goodfellow2014generative} to improve the perceptual realism of local image structures. The perceptual loss and adversarial loss, both computed on the image prediction $\hat{\bm{x}}_0^t = (\bm{x}_t - \sqrt{1-\alpha_t}\bm{\epsilon}_\theta(\bm{x}_t,\bm{y}, t))/\sqrt{\alpha_t}$, can be formulated as
\begin{eqnarray}
    \mathcal{L}_\textnormal{perc}&=& \mathbb{E}_{t,\bm{x}_0,\bm{\epsilon}}  \|\bm{\psi}_m(\hat{\bm{x}}_0^t) - \bm{\psi}_m(\bm{x}_0) \|,\\
    \mathcal{L}_\textnormal{adv}&=&   
    % \min_{\bm{\epsilon}_\theta}\max_{\delta_\theta}
    \mathbb{E}_{t,\bm{x}_0,\bm{\epsilon}} \left[\log D_{\theta}(\hat{\bm{x}}_0^t)\right]   +    \mathbb{E}_{\bm{x}_0} \left[\log (1 - D_{\theta}(\bm{x}_0) )\right],
\end{eqnarray}
where $D_{\theta}$ is the adversarial discriminator that tries to maximize $\mathcal{L}_{\textnormal{adv}}$, and $\bm{\psi}_m$ denotes the multilevel features from a pretrained VGG network.

\subsection{Normalized classifier-free guidance}
The diffusion model may ignore the conditional input and produce results uncorrelated with this input. One way to address this is the classifier-free guidance~\cite{ho2021classifier}, which considers $p(\bm{x}_t|\bm{y})$ along with $p(\bm{y} | \bm{x}_t)$ during sampling. The gradient of the log-probability $p(\bm{y} | \bm{x}_t)$ can be estimated as
\begin{equation}
  \begin{aligned}
    \nabla_{\bm{x}_{t}} \log p \left(\bm{y} | \bm{x}_{t}\right) & \propto \nabla_{\bm{x}_{t}} \log p\left(\bm{x}_{t} | \bm{y}\right)-\nabla_{\bm{x}_{t}} \log p\left(\bm{x}_{t}\right)  \propto \bm{\epsilon}_{\theta}\left(\bm{x}_{t} | \bm{y}\right)-\bm{\epsilon}_{\theta}\left(\bm{x}_{t}\right).
  \end{aligned}
\end{equation}
During sampling, we can estimate noise with a given condition $\bm{y}$ and a null condition $\bm{\emptyset}$ respectively, and generate samples further away from $\bm{\epsilon}_{\theta}(\bm{x}_t|\bm{\emptyset})$:
\begin{equation}
  \hat{\bm{\epsilon}}_{\theta}\left(\bm{x}_{t} | \bm{y}\right)=\bm{\epsilon}_{\theta}\left(\bm{x}_{t} |  \bm{y}\right)+ w \cdot\left(\bm{\epsilon}_{\theta}\left(\bm{x}_{t} | \bm{y}\right)-\bm{\epsilon}_{\theta}\left(\bm{x}_{t} | \bm{\emptyset}\right)\right),
  \label{eqn:classifierfree}
\end{equation}
where $w \geq 0$ controls the guidance strengths.
Such classifier-free guidance would trade off the sampling diversity to improve the quality of individual samples.

However, we observe that such a sampling procedure causes the mean and variance shift which hinders the subsequent denoising.
To be concrete, the guided noise sample $\hat{\bm{\epsilon}}_{\theta}\left(\bm{x}_{t} | \bm{y}\right)$ from  Equation~\ref{eqn:classifierfree} takes the mean as $\hat{\mu} = \mu + w (\mu - \mu_{\emptyset})$, indicating that there is a mean shift brought by the classifier-free guidance. Similarly, the variance of the noise sample shifts as $\hat{\sigma}^2 = (1 + w)^2\sigma^2 + w^2\sigma_{\emptyset}^2$ with an assumption that $\bm{\epsilon}_{\theta}\left(\bm{x}_{t} |  \bm{y}\right)$ and $\bm{\epsilon}_{\theta}\left(\bm{x}_{t} | \bm{\emptyset}\right)$ are independent variables. Such statistics shift will accumulate through all the $T$ diffusion denoising steps, leading to over-saturated images with over-smoothed textures.

To solve this, we propose normalized classifier-free guidance that explicitly matches the statistics of guided noise sample $\hat{\bm{\epsilon}}_{\theta}\left(\bm{x}_{t} | \bm{y} \right)$
according to the original estimation $\bm{\epsilon}_{\theta}\left(\bm{x}_{t} | \bm{y} \right)$, specifically,
\begin{equation}
    \tilde{\bm{\epsilon}}_\theta(\bm{x}_t|\bm{y}) = {\frac{\sigma}{\hat{\sigma}}} (\hat{\bm{\epsilon}}_\theta(\bm{x}_t|\bm{y}) - \hat{\mu} ) + \mu.
\end{equation}
We will show that the proposed normalized classifier-free guidance can effectively improve the sampling quality especially for large guidance strength $w$.

\section{Experiments}

\subsection{Implementation details}
 
We adopt a two-stage finetuning scheme. First, we fix the decoder and train the encoder with a learning rate of $3.5\mathrm{e}{-5}$ and a batch size of $128$. In the second stage, we train the full model jointly with a learning rate of $3\mathrm{e}{-5}$. We utilize AdamW optimizer~\cite{adamw} and also apply exponential moving average (EMA) with a rate of $0.9999$ during training. We sample the base model with $250$ diffusion steps and the upsample model with $27$ steps. All the experiments are performed on NVIDIA Tesla 32G-V100 GPUs. 

\subsection{Evaluation}

We conduct experiments on three different image-to-image translation tasks:
\begin{itemize}[leftmargin=*]
    \itemsep=0.4mm
    \item \emph{Mask-to-image synthesis.}   ADE20K~\cite{zhou2017scene} consists of 20K indoor and outdoor images  with 150 annotated semantic classes for training.  COCO~\cite{caesar2018coco} contains 120K training images with complex spatial context with 182 semantic classes, which is challenging for image synthesis.
    \item \emph{Sketch-to-image synthesis.} We extract sketches of images via HED~\cite{xie2015holistically} and then binarize extracted sketches. We evaluate our method on COCO-Stuff~\cite{caesar2018coco} and a proprietary dataset consisting of landscape images collected from Flickr with 50K training images and 2K test images.
    \item \emph{Geometry-to-image synthesis.} We use DIODE~\cite{vasiljevic2019diode} that contains 25K training images and 770 test images with dense depth and normal maps.  
\end{itemize}

We compare with three strong baselines: Pix2PixHD~\cite{wang2018high},  SPADE~\cite{park2019SPADE}, and OASIS~\cite{oasis}. To the best of our knowledge, diffusion models have not yet performed the above tasks. Hence we provide a diffusion model as the baseline that shares the same architecture as ours but is trained from scratch.

\begin{table}[t]
  \footnotesize
  \centering
  \caption{Comparison of FID on  various image translation tasks with the best score highlighted.}
  \label{tab:metric1}
  \begin{tabular}{@{} l @{}c @{\hspace{2.5mm}}  c @{\hspace{3mm}}  c @{\hspace{3mm}} c @{\hspace{3mm}} c @{\hspace{3mm}} c @{\hspace{3mm}} c  @{}}
    \toprule
    Method                        &&   ADE20K  & COCO (Mask)           & Flickr (Mask)                      &  COCO (Sketch) & Flickr (Sketch)  & DIODE \\
    \midrule % In-table horizontal line
    Pix2PixHD~\cite{wang2018high} &  &  35.3       & 37.5   &   26.1  & 27.1 &  16.8 & 18.2 \\
    SPADE~\cite{park2019SPADE}    &  &  18.9       & 15.0 &     17.4  & 48.9  & 29.5 & 17.0 \\
    OASIS~\cite{oasis}            &  &  14.8       & 8.8  &     10.5  & -   & -    & -  \\
    Ours (from scratch)                &  &  16.3       & 13.0  &    10.6  &  13.0  & 9.4  & 13.9 \\
    Ours                          &  &  \textbf{8.9}       & \textbf{5.2}   &    \textbf{6.1}   &  \textbf{8.8}  &  \textbf{6.0} & \textbf{11.5} \\
    \bottomrule
  \end{tabular}
\end{table}

\begin{figure*}[t]
  \centering
  \small
  \setlength\tabcolsep{1pt}
  \renewcommand{\arraystretch}{0.6}
  \includegraphics[width=\linewidth]{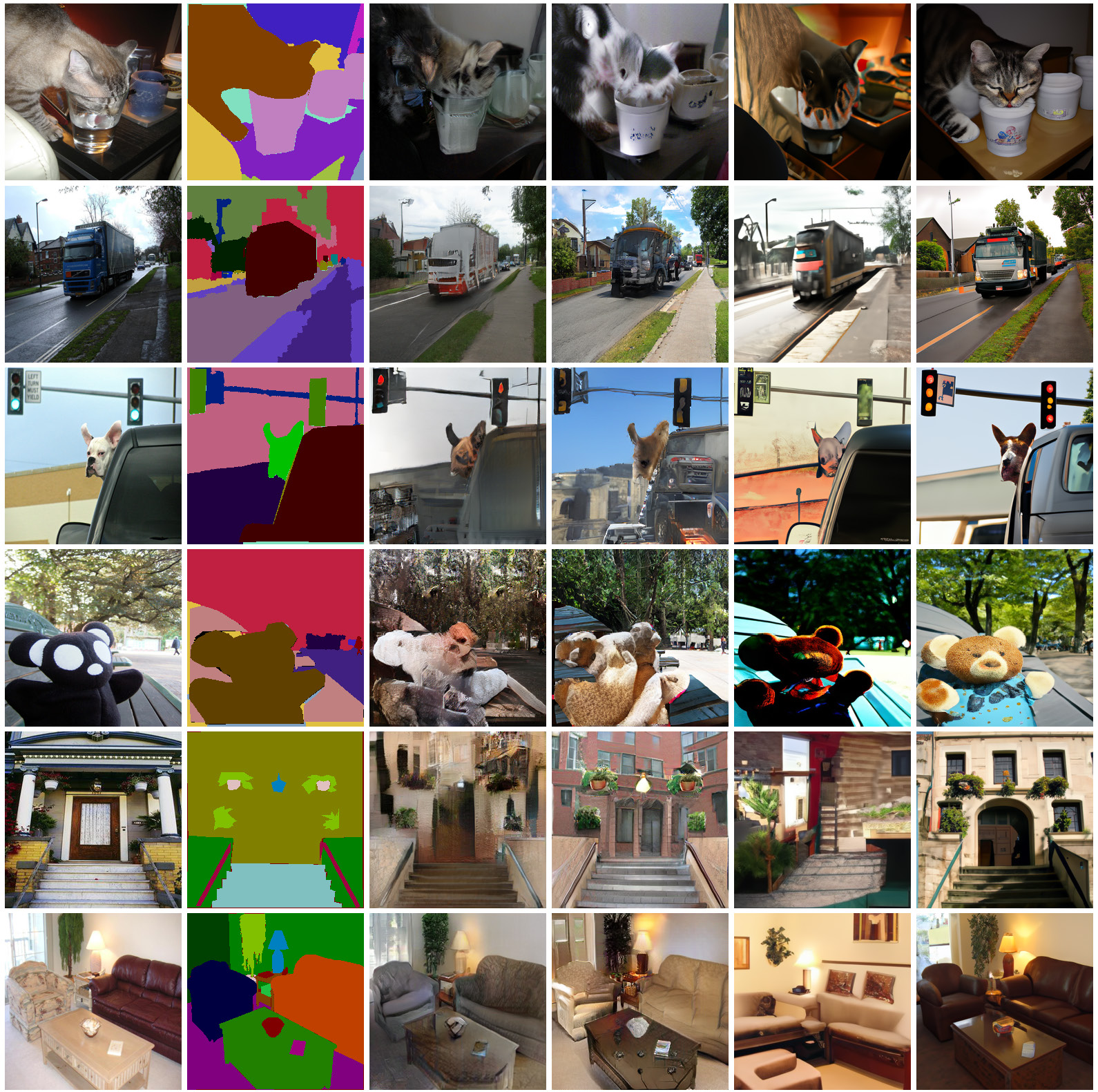}
    \begin{tabular}{@{} c@{\hspace{14mm}}c@{\hspace{12mm}}c@{\hspace{14mm}}c@{\hspace{10mm}}c@{\hspace{10mm}}c@{}}
    
    GT  & Condition & SPADE & OASIS & Ours (Scratch) & Ours
  \end{tabular}
  \caption{Visual comparisons on COCO and ADE20K. More results are shown in \textbf{Appendix}.}
  \label{fig:main1}
\end{figure*}

\begin{figure*}[t]
  \centering
  \small
  \setlength\tabcolsep{1pt}
  \renewcommand{\arraystretch}{0.6}
    \includegraphics[width=\linewidth]{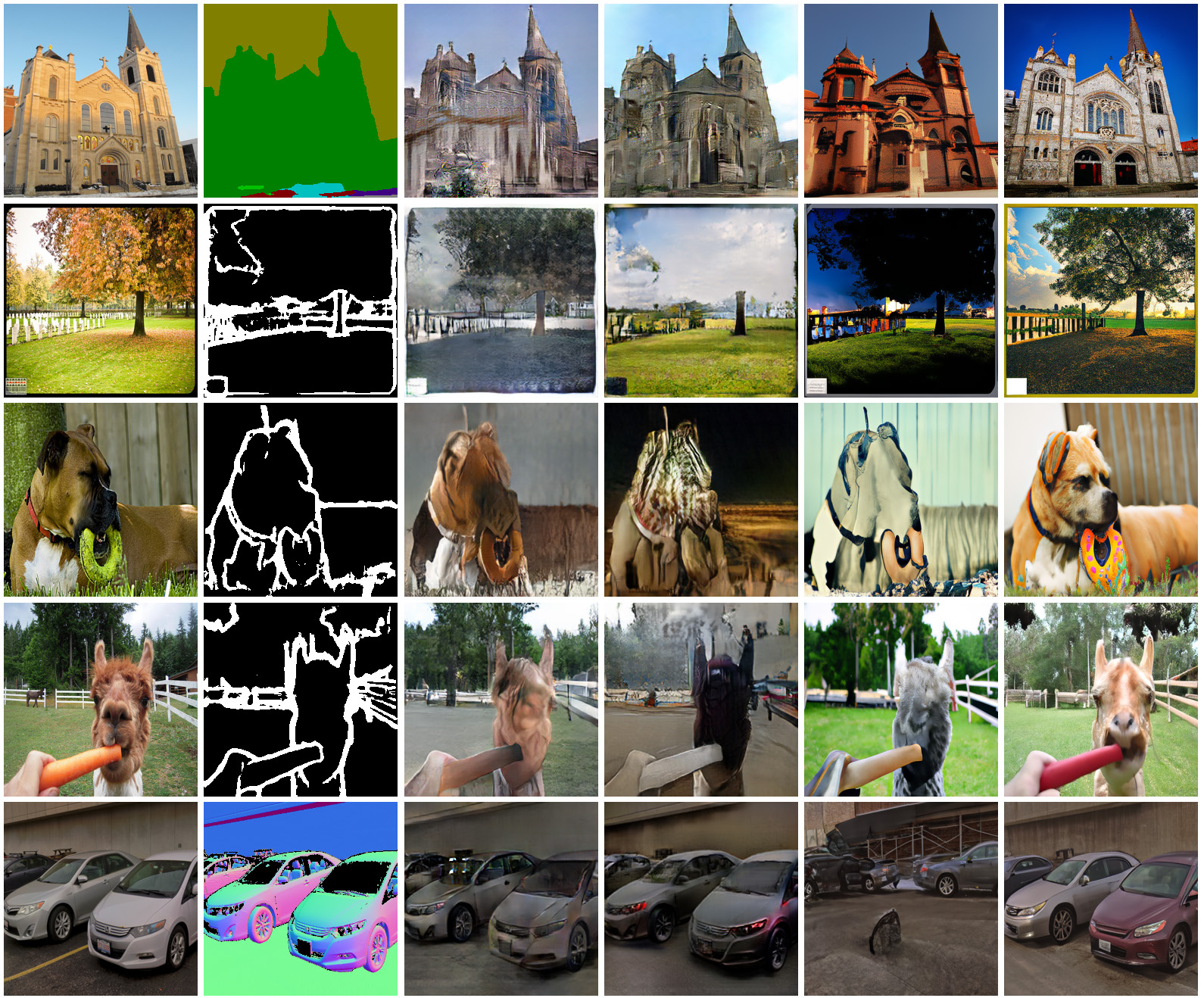}
    \begin{tabular}{@{\hspace{2mm}} c@{\hspace{12mm}}c@{\hspace{10mm}}c@{\hspace{10mm}}c@{\hspace{10mm}}c@{\hspace{10mm}}c@{}}

    GT    & Condition & Pix2PixHD & SPADE & Ours (Scratch) & Ours
  \end{tabular}
  \caption{Visual comparisons on other datasets. More results are shown in \textbf{Appendix}.}
  \label{fig:main2}
\end{figure*}

\begin{table}[t]
  \centering
  \footnotesize
  \caption{{User study on COCO. We report the preference rate of our approach over baselines.}}
  \label{tab:user}
  \renewcommand{\arraystretch}{1.1}
  \begin{tabular}{@{}l@{\hspace{6mm}}c@{\hspace{4mm}}c@{\hspace{4mm}}c@{}}
    \toprule        & Ours $>$ SPADE & Ours $>$ OASIS & Ours $>$  Scratch \\    \midrule
    Preference Rate & 93.6\%         & 84.1\%         & 87.4 \%           \\
    \bottomrule
  \end{tabular}
\end{table}

\begin{table*}[t]
\centering
\footnotesize
\caption{Ablation study of the proposed PITI on ADE20K dataset.}
\subfloat[{Finetune strategy}. 
\label{tab:abl1}
]{
\centering
\vspace{3.7mm}
\begin{minipage}{0.4\linewidth}{\begin{center}
    \tablestyle{4pt}{1.05}
    \begin{tabular}{@{}l@{\hspace{6mm}} c@{} }
      \toprule
      Finetune strategy       & FID  \\           
      \midrule
      Fixed decoder      & 12.6  \\
      One-stage finetune & 13.3\\
      Two-stage finetune & 8.9 \\
      
      \bottomrule
    %   \label{tab:abl1}
    \end{tabular}
\end{center}}\end{minipage}
}
\hspace{1em}
\subfloat[{Upsampling strategy}. 
\label{tab:sr}
]{
\begin{minipage}{0.5\linewidth}{\begin{center}
    \tablestyle{4pt}{1.05}
    \begin{tabular}{@{} ccc@{\hspace{6mm}}c@{} }
      \toprule
      
      Degradation & $\mathcal{L}_{\textnormal{perceptual}}$ & $\mathcal{L}_{\textnormal{adversarial}}$  & FID  \\
      \midrule
                  &                 &            & 14.5 \\
      \checkmark  &                 &            & 12.1 \\
      \checkmark  & \checkmark      &            & 9.8  \\
      \checkmark  & \checkmark      & \checkmark & 8.9  \\
      \bottomrule
    %   \label{tab:sr}
    \end{tabular}
\end{center}}\end{minipage}
}
\label{tab:ablations}
\end{table*}

\begin{table}[ht]
    \centering
    \small
     \caption{Comparison on FID with different training image sizes.}
    \label{tab:few-shot}
    \begin{tabular}{cccccc}
    \toprule
        Training Size & Pix2PixHD & SPADE & OASIS & Ours (From Scratch) & Ours \\
    \midrule
      25\%  & 44.1 & 27.1 & 26.5 & 24.0& \textbf{16.2}\\
      50\%  & 38.4 & 24.6 & 20.4 & 18.8& \textbf{12.7}\\
      100\%  &35.3 & 18.9& 14.8& 16.3& \textbf{8.9}\\
    \bottomrule
    \end{tabular}
\end{table}

\paragraph{Quantitative results.}
Recent work~\cite{kynkaanniemi2022role} finds that FID measured by InceptionNet~\cite{heusel2017gans} may not correlate with the perceptual quality, as the model is initially trained for ImageNet classification. Following~\cite{kynkaanniemi2022role},   we calculate FID~\cite{heusel2017gans} with the CLIP model~\cite{radford2021learning} whose feature space is more robust and transferable. Table~\ref{tab:metric1} shows our method consistently outperforms the model without pretraining by a large margin. Compared with the leading approach, OASIS~\cite{oasis}, on mask-to-image synthesis, our method obtains significant improvements ($5.9$ on ADE20K, $3.6$ on COCO, and $4.4$ on Flickr) in terms of FID. Our general approach also shows promising performance on sketch-to-image and geometry-to-image synthesis tasks. 

\paragraph{Qualitative results.}
We show visual results of different tasks in Figure~\ref{fig:main1} and Figure~\ref{fig:main2}. Compared with from-scratch methods that suffer severe artifacts for complex scenes, the pretrained model significantly improves the quality and diversity of generated images. As the COCO dataset contains many categories with diverse combinations, all the baseline methods fail to generate visually-pleasing results with compelling structures. In contrast, our methods can produce vivid details with correct semantics even for challenging cases. Figure~\ref{fig:main2} shows good applicability of our approach to different input modalities. 

\paragraph{Human evaluation.} We also perform a user study on mask-to-image synthesis on COCO-Stuff on the Amazon Mechanical Turk,  with 3,000 votes from 20 participants. Participants are given a pair of images at once and are asked to select a more realistic one. As shown in Table~\ref{tab:user}, the proposed approach outperforms from-scratch models and other baselines by a large margin.

\subsection{Ablation study}
\label{sec:ablation}
\begin{figure*}[t]
  \centering
  \small
     \includegraphics[width=\linewidth]{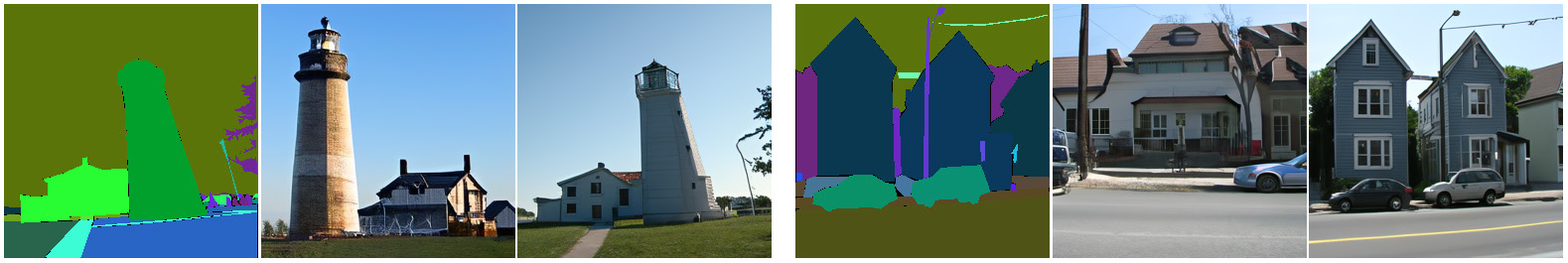}
  \begin{tabular}{@{\hspace{4mm}} c@{\hspace{10mm}}c@{\hspace{8mm}}c@{\hspace{13mm}}c@{\hspace{10mm}}c@{\hspace{8mm}}c@{}}
   
      Input & Fixed decoder & Full model &  Input & Fixed decoder & Full model\\
 
  \end{tabular}
  \caption{Fixing the decoder generates high-quality images but fails to align with the condition. }
  \label{fig:fixed-decoder}
\end{figure*}

\paragraph{Effect of two-stage finetune strategy.}
To analyze the importance of the finetune strategy, we perform three finetune schemes on ADE20K: (a) \textbf{Fixed decoder} where we freeze the pretrained decoder and only train a task-specific encoder, (b) \textbf{One-stage finetune} where we finetune both encoder and the pretrained decoder simultaneously, and (c) \textbf{Two-stage finetune} where we first train the encoder with decoder fixed, and then finetune them jointly. As shown in Table~\ref{tab:abl1}, the proposed two-stage finetune pipeline achieves the best performance. Interestingly, we observe that with the decoder fixed, the generated images are of high visual quality but fail to align with the given semantic map, as shown in Fig.~\ref{fig:fixed-decoder}. This demonstrates that the pretrained decoder has the generative prior to synthesizing a realistic image from a latent semantic vector.

\paragraph{Adversarial diffusion upsampler.}
We propose to improve the upsampling diffusion model with  degradations on input images and image-level losses on the output noises. Table~\ref{tab:sr}  gives the quantitative comparison of different upsampling settings. We can see that enforcing degradations on the input leads to better FID scores. On the other hand, with reconstruction loss on the predicted noises only, the upsampled images contains fewer high-frequency details. In contrast, the perceptual and adversarial losses can significantly improve the upsampling quality.

\begin{figure*}[t]
  \centering
  \small
     \includegraphics[width=\linewidth]{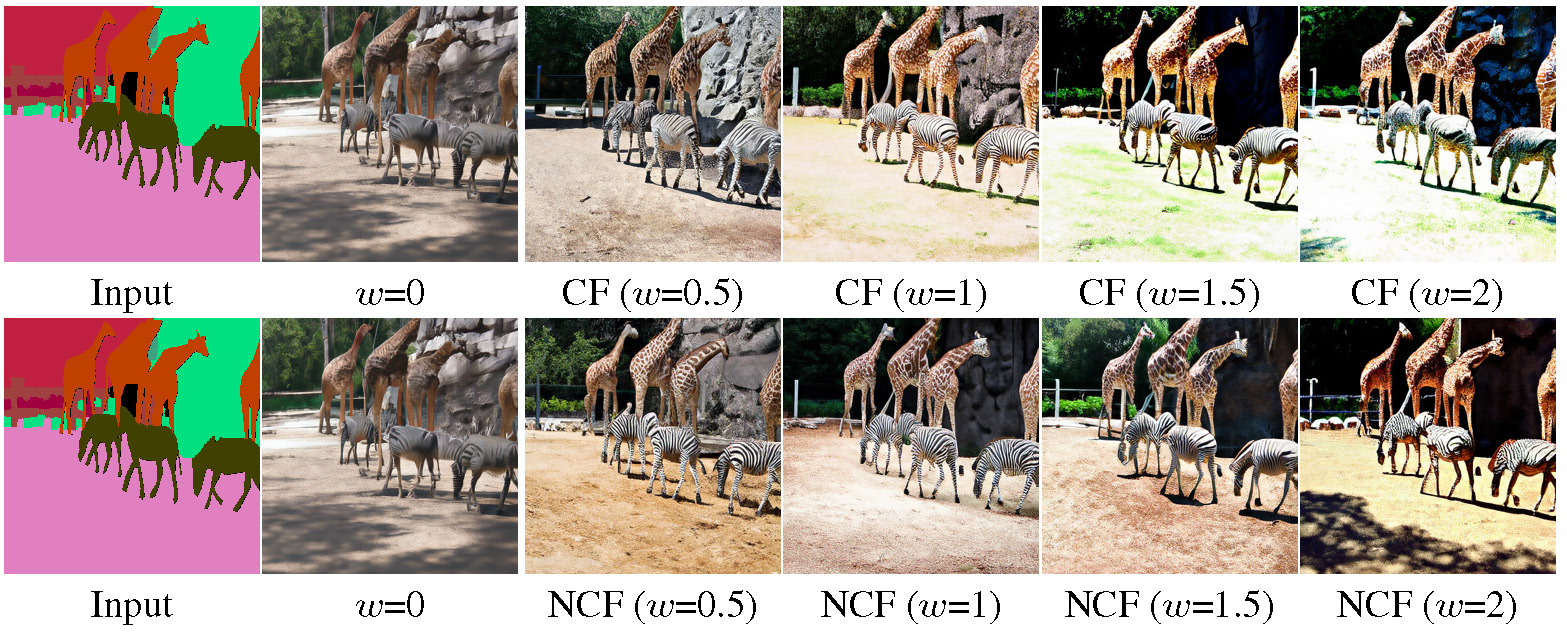}
  \caption{Effect of normalized classifier-free (NCF) guidance sampling. }
  \label{fig:guidance}
\end{figure*}

\paragraph{Normalized classifier-free guidance.}
Figure~\ref{fig:guidance} compares  the normalized classifier-free guidance (NCF) against the original classifier-free guidance (CF) with different guidance strengths $w$. When sampling with a large $w$, CF tends to produce smooth content without much detailed structures  while the sampled results using NCF exhibits more vivid details.

\paragraph{Smaller training dataset.} 
Image synthesis tasks often suffer from limited training data, which hinders training a high-quality generative model from scratch. To show how the pretraining alleviates the needs of data, we reduce the number of training images of ADE20K to 25\% (5k) and 50\% (10k), respectively, and report the FID in Table~\ref{tab:few-shot}. With only 25\% of training data, the proposed method achieves a comparable FID to previous methods trained on full data.

\section{Conclusion and Limitation}
We present a simple and universal framework that brings the power of pretraining to various image-to-image translation tasks. Enhanced by techniques like adversarial diffusion upsampler and normalized classifier-free guidance, the full model, PITI significantly advances the  state-of-the-art synthesis quality especially in challenging scenarios. 
One limitation of our method is that the sampled images have difficulties in faithfully aligning with the given inputs and may miss the small objects. One possible  reason is that the intermediate space of the pretrained model lacks accurate spatial information. We plan to explore other ways for pretraining in the future.  We hope the work can inspire more works along the path and advance the field towards realistic synthesis.

\section*{Broader Impact}
Conditional image synthesis aims at generating high-quality images with faithfulness to the given condition. It is a fundamental problem in computer vision and graphics, enabling diverse content creation and manipulation.
Large-scale pretraining has shown superior performance on high-level vision such as image classification, object detection, and semantic segmentation. However, whether general generative tasks can also benefit from large-scale pretraining remains unanswered. This work thus explores the role of pretraining in conditional image synthesis and demonstrates that image pretraining can significantly improve the generation quality with the proposed finetuning method for complex objects and general scenes. Our work bridges the gap between image pretraining and image synthesis and provides new insight into image-to-image translation.

The main concerns of the image pretraining are energy consumption and carbon emission. Though the pretraining is energy-consuming, we only need to train the model once.  Different downstream tasks can then share the same pretrained model via conditional finetuning. Also, the pretraining enables training a generative model with fewer training data, thereby advancing the progress of image synthesis when data is limited due to privacy issues or high annotation cost.

{\small
  \bibliographystyle{plain}
  \bibliography{ref.bib}
}

\clearpage

\appendix
\section*{Appendix}

\section{Implementation details}
Our decoder is inherited from GLIDE~\cite{nichol2021glide}, which adopts the model architecture proposed in a recent work~\cite{dhariwal2021diffusion}. To adapt the pretrained diffusion model to image-to-image translation,  we design a condition encoder to map  input conditions to semantic tokens, which combines convolutional layers and ViT~\cite{vit}. We set the diffusion steps for both the base model and upsampling model as 1,000.  

We adopt a two-stage finetuning scheme. First, we fix the decoder and train the encoder for 200K iterations with a learning rate of $3.5\mathrm{e}{-5}$ and a batch size of $128$.   In the second stage, we finetune the full model jointly with a learning rate of $3\mathrm{e}{-5}$.   We utilize AdamW optimizer~\cite{adamw} and apply exponential moving average (EMA) with a rate of $0.9999$ during training. We sample the base model with $250$ diffusion steps and the upsample model with $27$ steps.

\section{Additional results}
\begin{figure*}[t]
    \centering 
    \small
    \includegraphics[width=\linewidth]{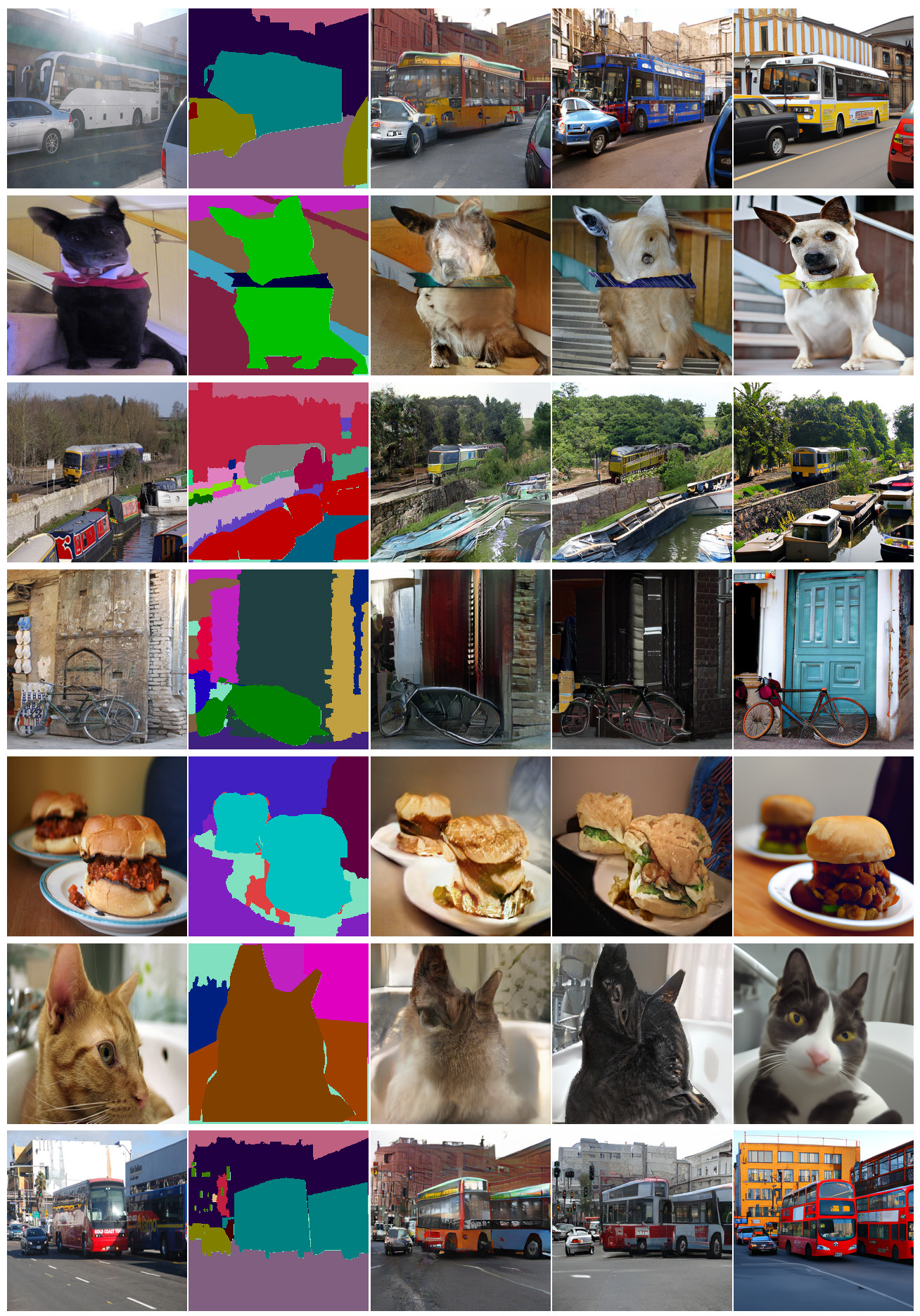} \\ 
    \begin{tabular}{@{} c@{\hspace{18mm}}c@{\hspace{18mm}}c@{\hspace{18mm}}c@{\hspace{18mm}}c@{}}
     GT&   Condition &SPADE&OASIS &Ours 
    \end{tabular}
   \caption{Visual comparisons on mask-to-image synthesis on COCO.}
    \label{fig:coco-mask1}
\end{figure*}

\begin{figure*}[t]
    \centering 
    \small
    \includegraphics[width=\linewidth]{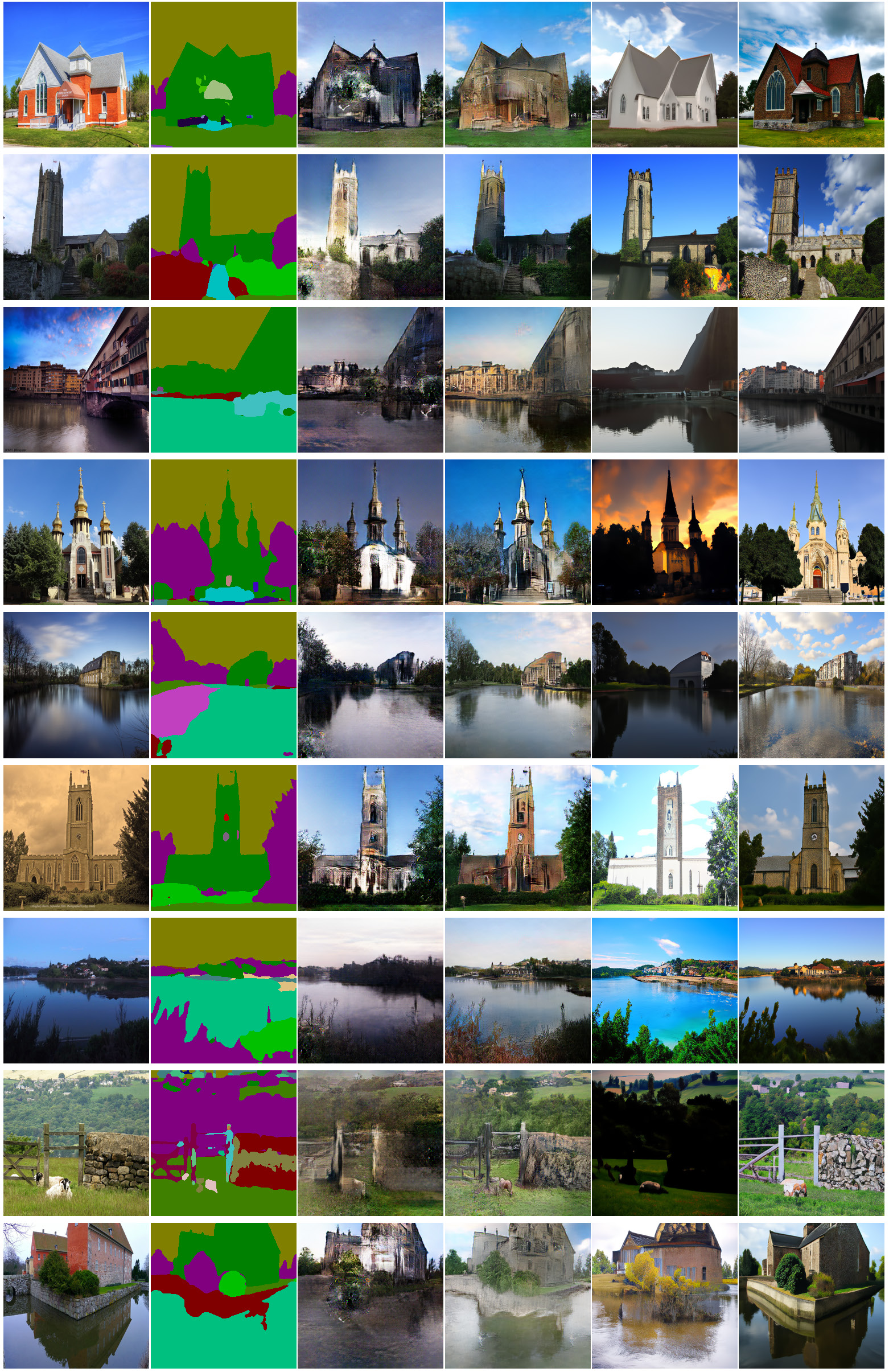} \\ 
    \begin{tabular}{@{} c@{\hspace{14mm}}c@{\hspace{10mm}}c@{\hspace{12mm}}c@{\hspace{9mm}}c@{\hspace{10mm}}c@{}}
  
     GT&   Condition & Pix2PixHD & Spade &From Scratch &Ours       
    \end{tabular}
 
   \caption{Visual comparisons on mask-to-image synthesis on Flickr.}
    \label{fig:flk-mask}
\end{figure*}

\begin{figure*}[t]
    \centering 
    \small
    \includegraphics[width=0.9\linewidth]{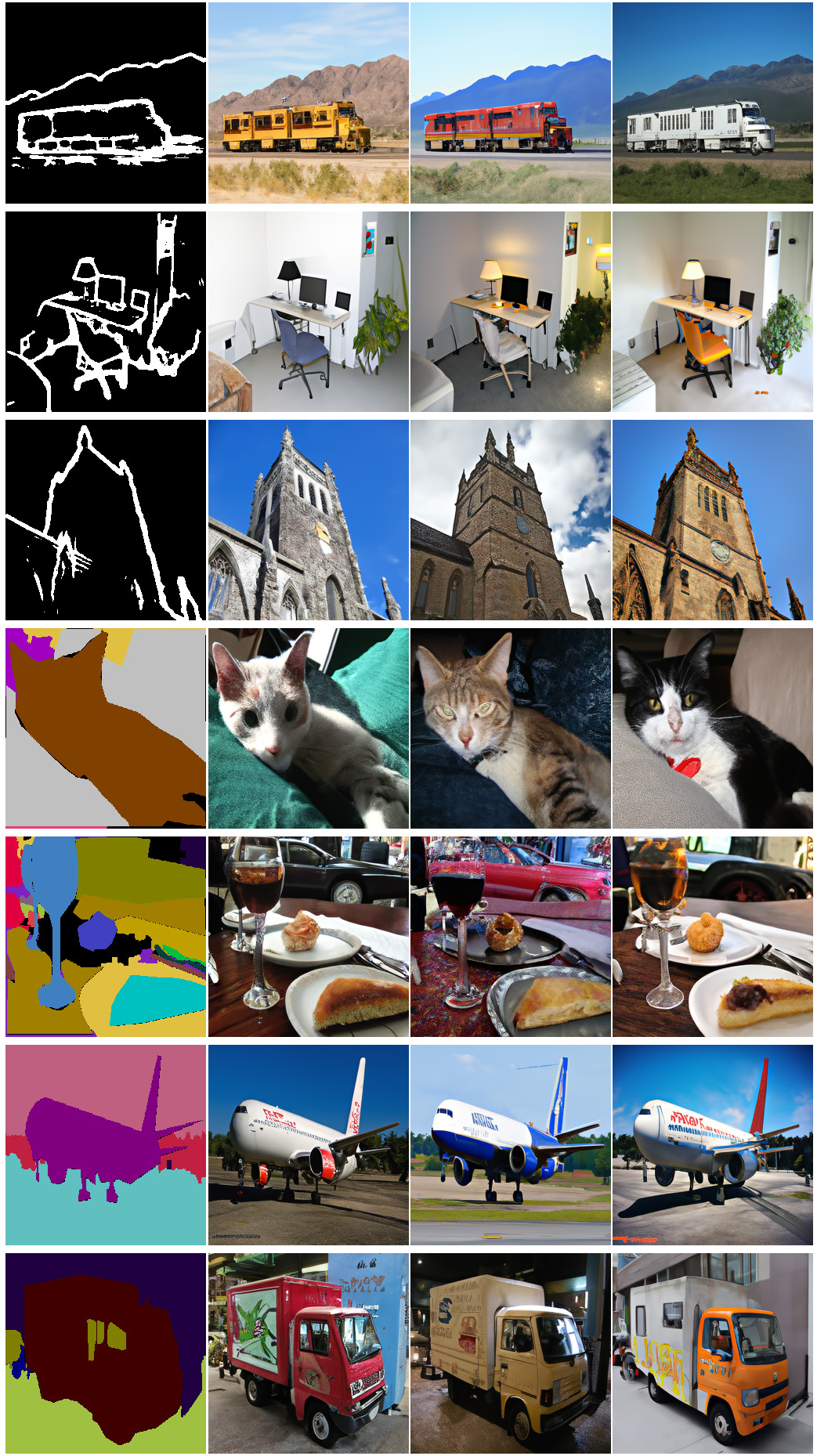} \\ 
    \begin{tabular}{@{\hspace{4mm}} c@{\hspace{22mm}}c@{\hspace{20mm}}c@{\hspace{20mm}}c@{}}
   
      Input&   Sample 1 &Sample 2&Sample 3   
    \end{tabular}
   \caption{Diverse samples from our model.}
    \label{fig:diverse}
\end{figure*}

\textbf{Visual results.}
We present additional visual results compared with previous approaches in Fig.~\ref{fig:coco-mask1}$\sim$~\ref{fig:flk-mask}.
% Fig.~\ref{fig:coco-mask2}, Fig.~\ref{fig:coco-mask3}, Fig.~\ref{fig:coco-mask4}, Fig.~\ref{fig:coco-edge1}, Fig.~\ref{fig:coco-edge2}, Fig.~\ref{fig:ade-mask1},Fig.~\ref{fig:ade-mask2}, Fig.~\ref{fig:ade-mask3}, Fig.~\ref{fig:diode_1}, Fig.~\ref{fig:flk1}, and Fig.~\ref{fig:flk-mask}. 
We also demonstrate diverse plausible outputs sampled by our model in Fig.~\ref{fig:diverse}, which shows the generated images pose both high quality and diversity.

\begin{figure*}[t]
    \centering 
    \small
    \includegraphics[width=0.88\linewidth]{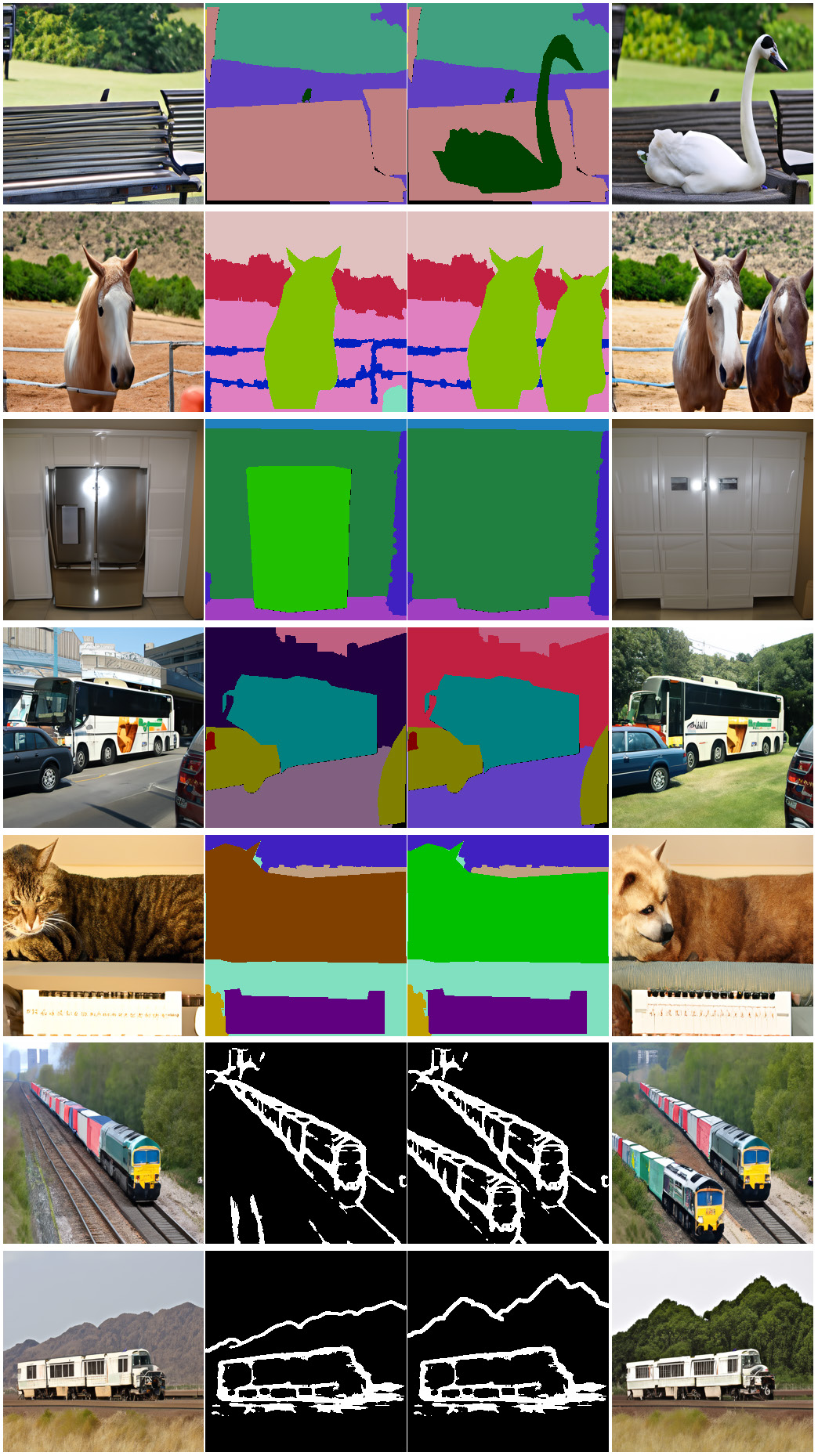} \\ 
    \begin{tabular}{@{\hspace{2mm}} c@{\hspace{9mm}}c@{\hspace{8mm}}c@{\hspace{12mm}}c@{}}
   
      Original image&   Original condition & Edited condition& Edited image   \\
    \end{tabular}
   \caption{Results of image editing, such as image composition, object removal, change of semantic class, and change of shape.}
    \label{fig:edit}
\end{figure*}

\textbf{Image manipulation.}
Fig~\ref{fig:edit} shows  the proposed model can be used for various image manipulation, such as image composition, object removal, change of semantic class, and change of shape. To preserve the content of unedited regions in original images, we replace the DDPM sampling procedure with DDIM~\cite{song2020denoising} where samples are uniquely determined from given noise latent variables.

\textbf{Numerical results.}
 Many recent works~\cite{kynkaanniemi2022role,ramesh2022hierarchical} find that FID measured by InceptionNet~\cite{heusel2017gans}  does not always align with the human judge on perceptual quality, as the model is initially trained for ImageNet classification. In contrast, CLIP model~\cite{radford2021learning}   is more robust and transferable for FID evaluation. To quantitatively evaluate the generation capacity of different generative models, we report FID calculated with CLIP (FID-C) in Table~\ref{tab:metric2}. We also report FID calculated by InceptionNet (FID-I) for reference.

 \begin{table}[h]
  \scriptsize
  \centering
  \caption{Quantitative comparison on  diverse image translation tasks.}
  \label{tab:metric2}
  \begin{tabular}{@{} l @{}  c @{\hspace{4mm}}          c @{\hspace{2mm}}  c @{\hspace{0.1mm}} c @{\hspace{2mm}} c @{\hspace{0.1mm}} c @{\hspace{0.1mm}}c@{\hspace{2mm}}  c @{\hspace{0.1mm}} c @{\hspace{0.1mm}} c @{\hspace{2mm}} c@{\hspace{0.1mm}}  c @{\hspace{0.1mm}} c @{\hspace{2mm}} c@{\hspace{0.1mm}}  c @{\hspace{0.1mm}} c @{\hspace{2mm}}  c @{\hspace{2mm}}   c @{}}
    \toprule
                                  &  & \multicolumn{2}{c}{ADE20K} &      & \multicolumn{2}{c}{ COCO (Mask)} &  &
                                  \multicolumn{2}{c}{ Flickr (Mask)} &  &\multicolumn{2}{c}{COCO (Sketch) } &      & \multicolumn{2}{c}{ Flickr (Sketch)} &  & \multicolumn{2}{c}{ DIODE}                                          \\
    \cmidrule{3-4} \cmidrule{6-7}   \cmidrule{9-10}  \cmidrule{12-13}  \cmidrule{15-16} \cmidrule{18-19}
    Method                        &  & FID-I                        & FID-C                        &  & FID-I                                       & FID-C          &  & FID-I   & FID-C                    &  & FID-I                        & FID-C    && FID-I  & FID-C &  & FID-I  & FID-C \\
    \midrule % In-table horizontal line
    Pix2PixHD  &  & 61.8                       & 35.3                             &  & 67.7                                 & 37.5          &  & 41.5     & 26.1        & & 38.7                        & 27.1    & & 26.9   & 16.8   &  & 66.0 & 18.2 \\
    SPADE     &  & 33.9                       & 18.9                             &  & 22.6                                 & 15.0          &  & 27.7     & 17.4        & & 89.2                        & 48.9    & & 43.6   & 29.5   &  & 61.2 & 17.0 \\
    OASIS           &  & 28.3                       & 14.8                             &  & 17.0                                 & 8.8           &  & 24.4     & 10.5        &  & -                          & -      & & -       & -       &  & -    & -    \\
    Ours (from scratch)                &  & 35.7                      & 16.3                              &  & 25.1                                 & 13.0     &  & 26.9     & 10.6       & & 33.6                        & 13.0    &  & 24.8  & 9.4     &  & 70.2 & 13.9 \\
    Ours                          &  & \textbf{27.3}                       & \textbf{8.9  }                            &  & \textbf{15.8 }                                & \textbf{5.2 }          &  & \textbf{21.2}     & \textbf{6.1 }     & & \textbf{21.4}                        & \textbf{8.8 }    &  & \textbf{20.3}  & \textbf{6.0 }    &  & \textbf{59.6} & \textbf{11.5} \\
    \bottomrule
  \end{tabular}
\end{table}

\section{Limitation}
\begin{figure*}[t]
    \centering 
    \small
    \includegraphics[width=\linewidth]{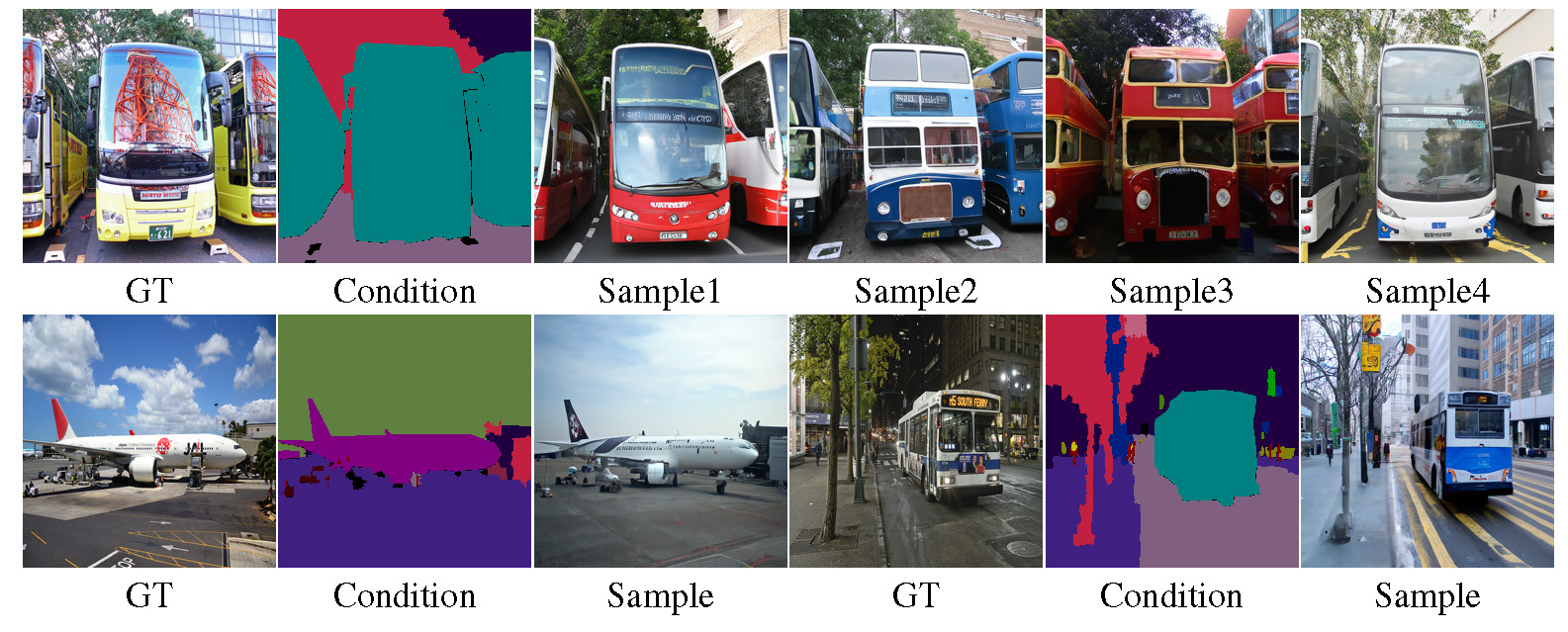} \\  
   \caption{Limitation. In the first row, though our model can produce diverse buses in different samples, we found buses within a single sample tend to exhibit similar style and color. Another limitation of the proposed method is that the samples do not always perfectly align with the input conditions, with some small objects missed. In the left example of the second row, trees in front of the airplane are not synthesized properly. Similarly, the lighting in the right example is missed.}
    \label{fig:limitation}
\end{figure*}

Fig~\ref{fig:limitation} presents the limitation of the proposed approach.  In the first row, we found that similar objects within a single sample are prone to highly-correlated styles, though our model can produce diverse samples of different styles. Distinguished from the well-known mode collapse issue where samples lack inter-image diversity, we call this intra-image mode collapse. Another limitation of our method is that the generated images are subject to minor misalignment with input conditions for some small objects. We conjecture that the latent space of our model lacks accurate spatial information, as the model is initially pretrained on text-to-image synthesis. We leave the exploration of other image pretraining manners as the future work.

\begin{figure*}[t]
    \centering 
    \vspace{-7mm}
    \small
    \includegraphics[width=\linewidth]{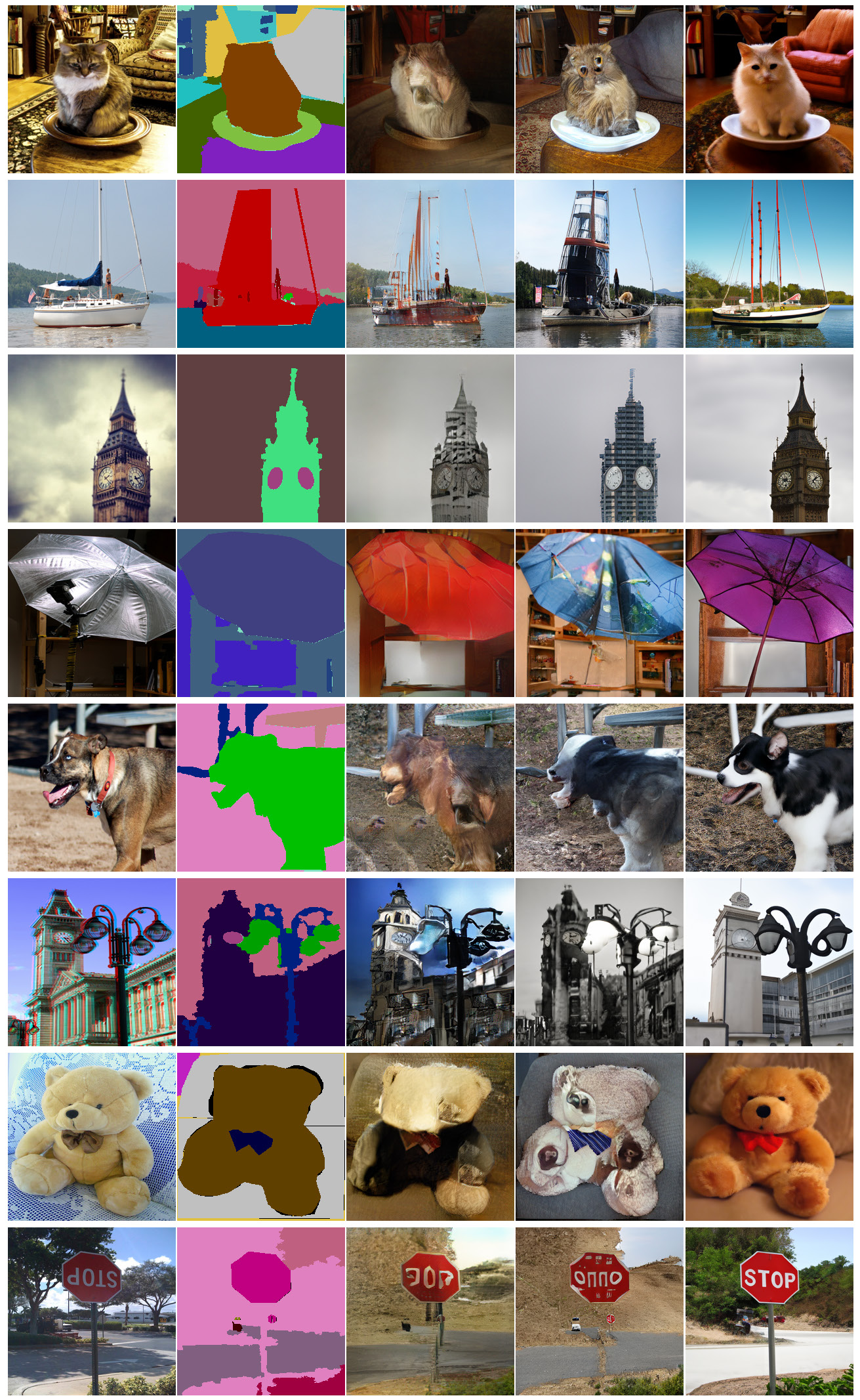} \\ 
    \begin{tabular}{@{} c@{\hspace{18mm}}c@{\hspace{18mm}}c@{\hspace{18mm}}c@{\hspace{18mm}}c@{}}
     
     GT&   Condition &SPADE&OASIS &Ours    
     
    \end{tabular}
   \caption{Visual comparisons on mask-to-image synthesis on COCO.}
    \label{fig:coco-mask2}
\end{figure*}    

\begin{figure*}[t]
 \vspace{-7mm}
    \centering 
    \small
    \includegraphics[width=\linewidth]{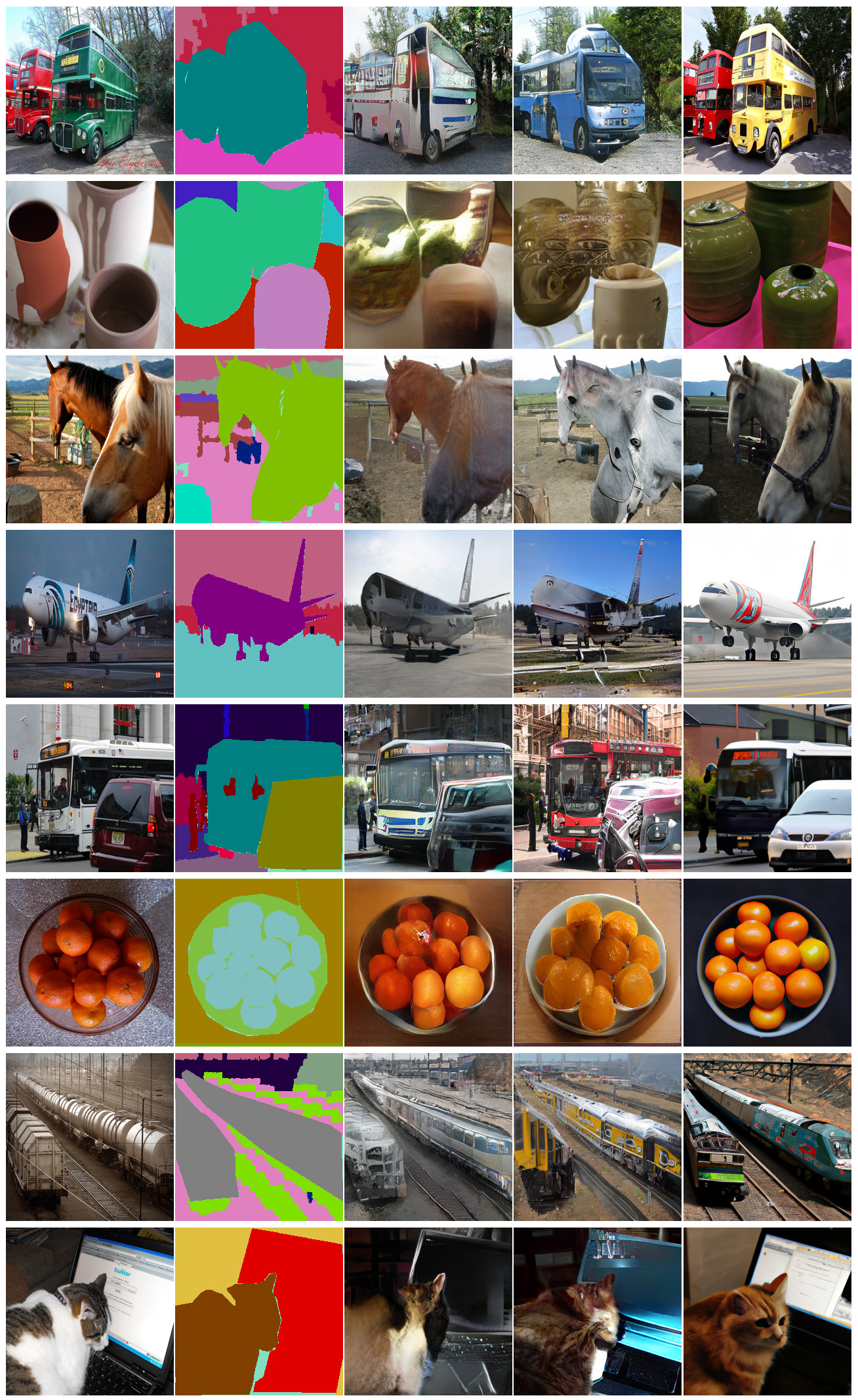} \\ 
    \begin{tabular}{@{} c@{\hspace{18mm}}c@{\hspace{18mm}}c@{\hspace{18mm}}c@{\hspace{18mm}}c@{}}
     GT&   Condition &SPADE&OASIS &Ours    
    \end{tabular}
   \caption{Visual comparisons on mask-to-image synthesis on COCO.}
    \label{fig:coco-mask3}
\end{figure*}    

\begin{figure*}[t]
\vspace{-7mm}
    \centering 
    \small
    \includegraphics[width=\linewidth]{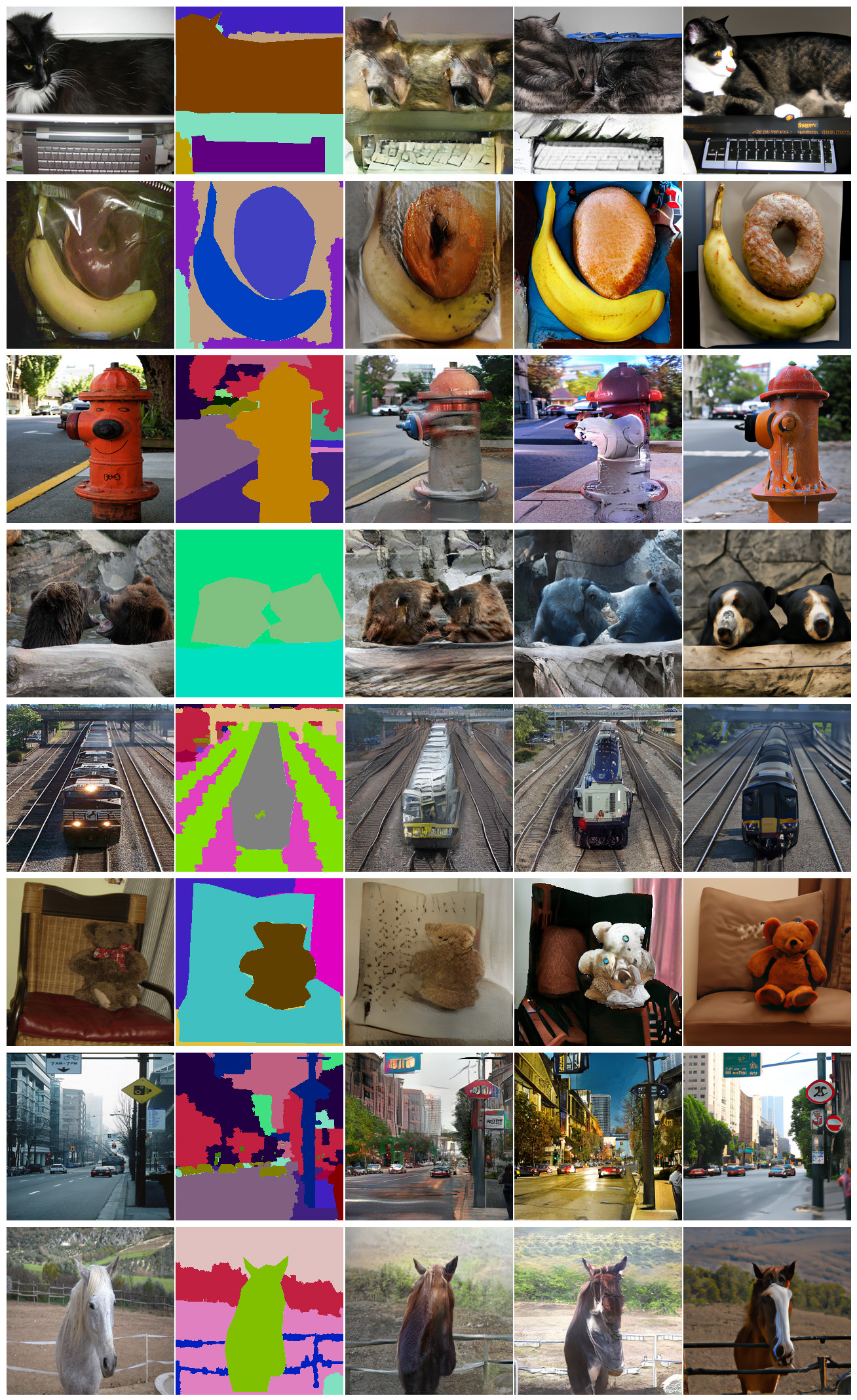} \\ 
    \begin{tabular}{@{} c@{\hspace{18mm}}c@{\hspace{18mm}}c@{\hspace{18mm}}c@{\hspace{18mm}}c@{}}

     GT&   Condition &SPASE&OASIS &Ours    
    \end{tabular}
 
   \caption{Visual comparisons on mask-to-image synthesis on COCO.}
    \label{fig:coco-mask4}
\end{figure*}

\begin{figure*}[t]
    \centering 
    \small
    \includegraphics[width=\linewidth]{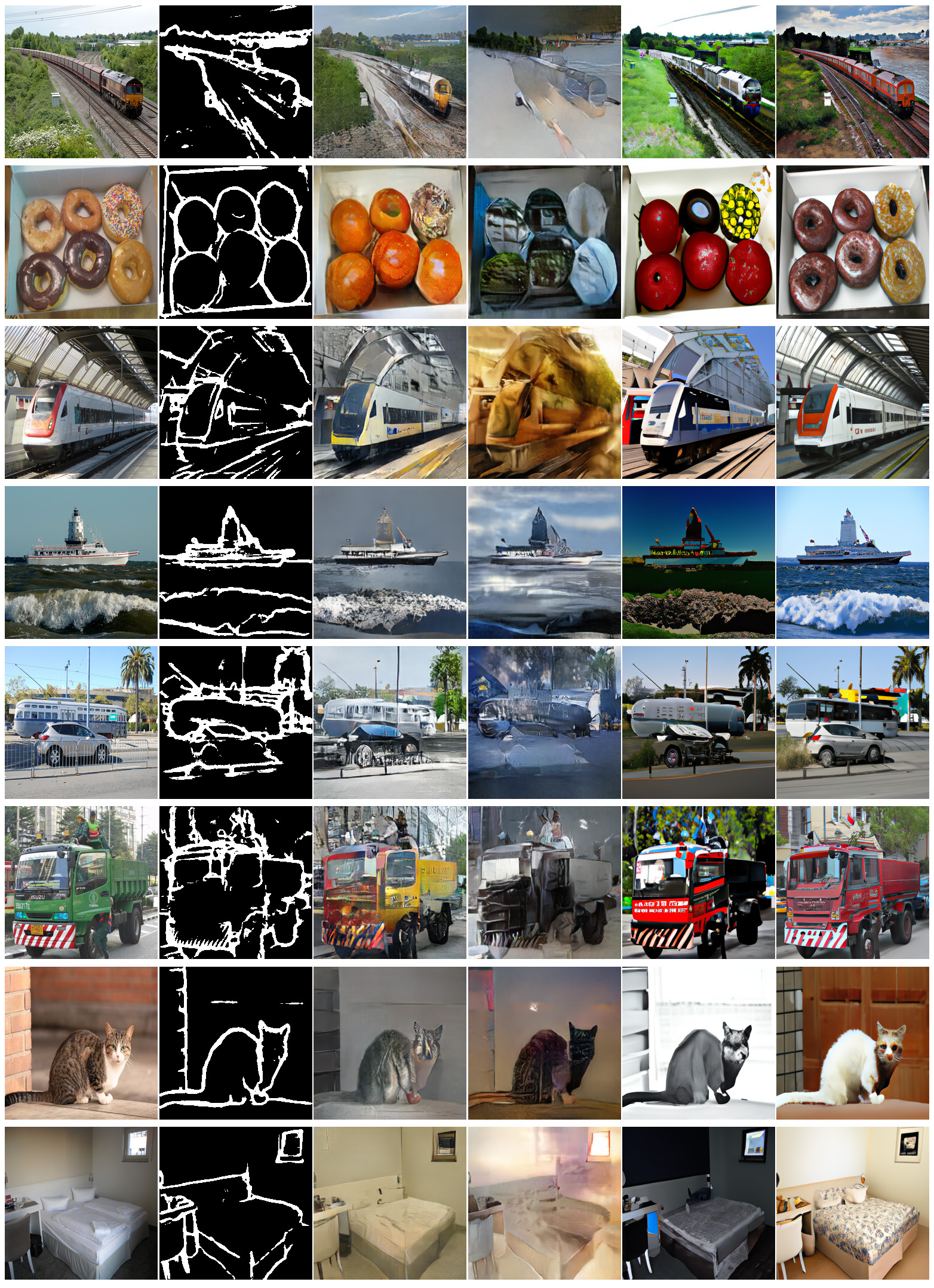} \\ 
     \begin{tabular}{@{} c@{\hspace{14mm}}c@{\hspace{10mm}}c@{\hspace{12mm}}c@{\hspace{9mm}}c@{\hspace{10mm}}c@{}}
                 
     GT&   Condition &Pix2PixHD &SPADE &From Scratch &Ours    
    \end{tabular}
 
   \caption{Visual comparisons on sketch-to-image synthesis on COCO.}
    \label{fig:coco-edge1}
\end{figure*}    

\begin{figure*}[t]
    \centering 
    \small
    \includegraphics[width=\linewidth]{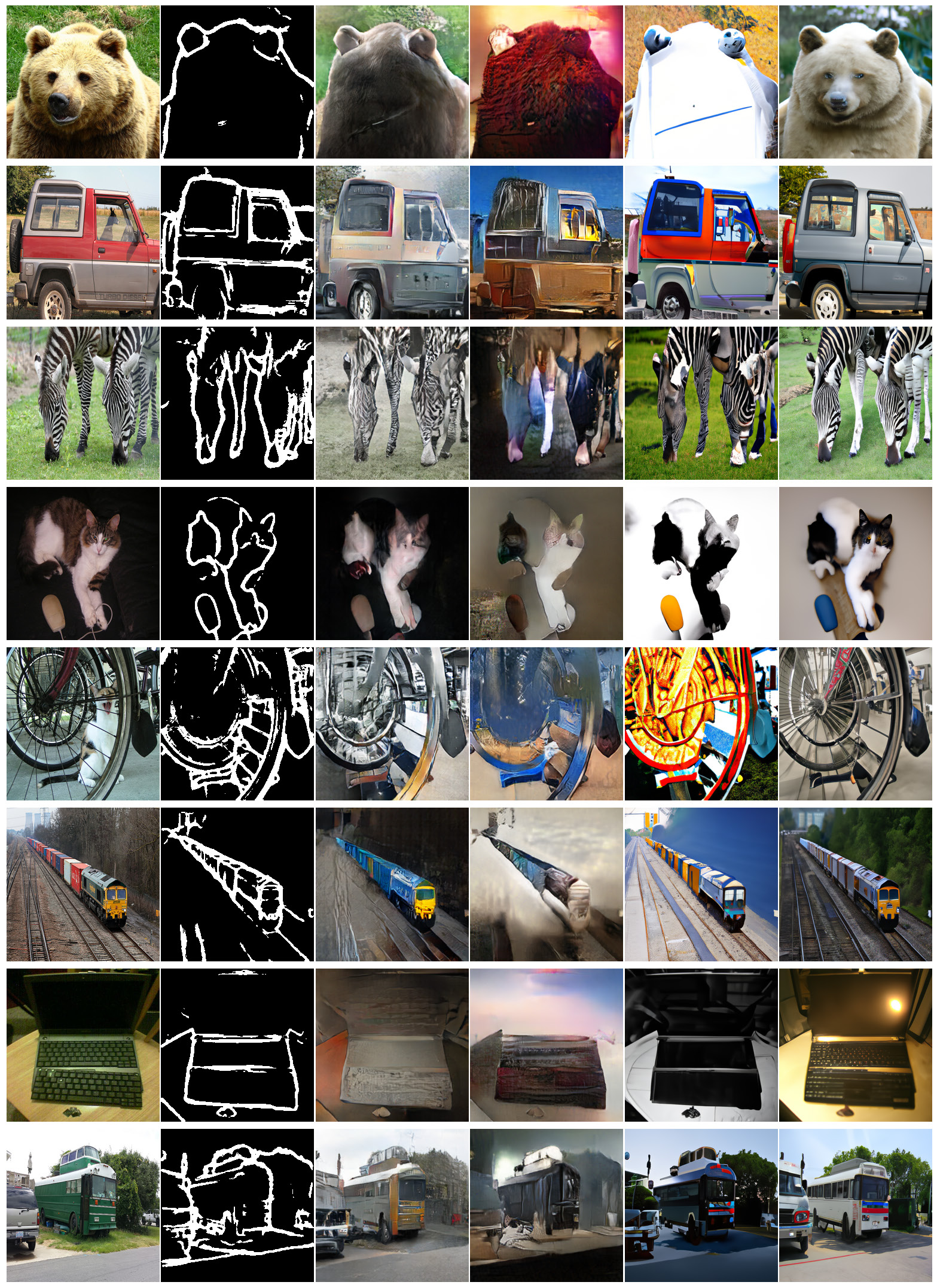} \\ 
    \begin{tabular}{@{} c@{\hspace{14mm}}c@{\hspace{10mm}}c@{\hspace{12mm}}c@{\hspace{9mm}}c@{\hspace{10mm}}c@{}}

     GT&   Condition &Pix2PixHD &SPADE &From Scratch &Ours    
    \end{tabular}
   \caption{Visual comparisons on sketch-to-image synthesis on COCO.}
    \label{fig:coco-edge2}
\end{figure*}

\begin{figure*}[t]
    \centering 
    \small
    \includegraphics[width=\linewidth]{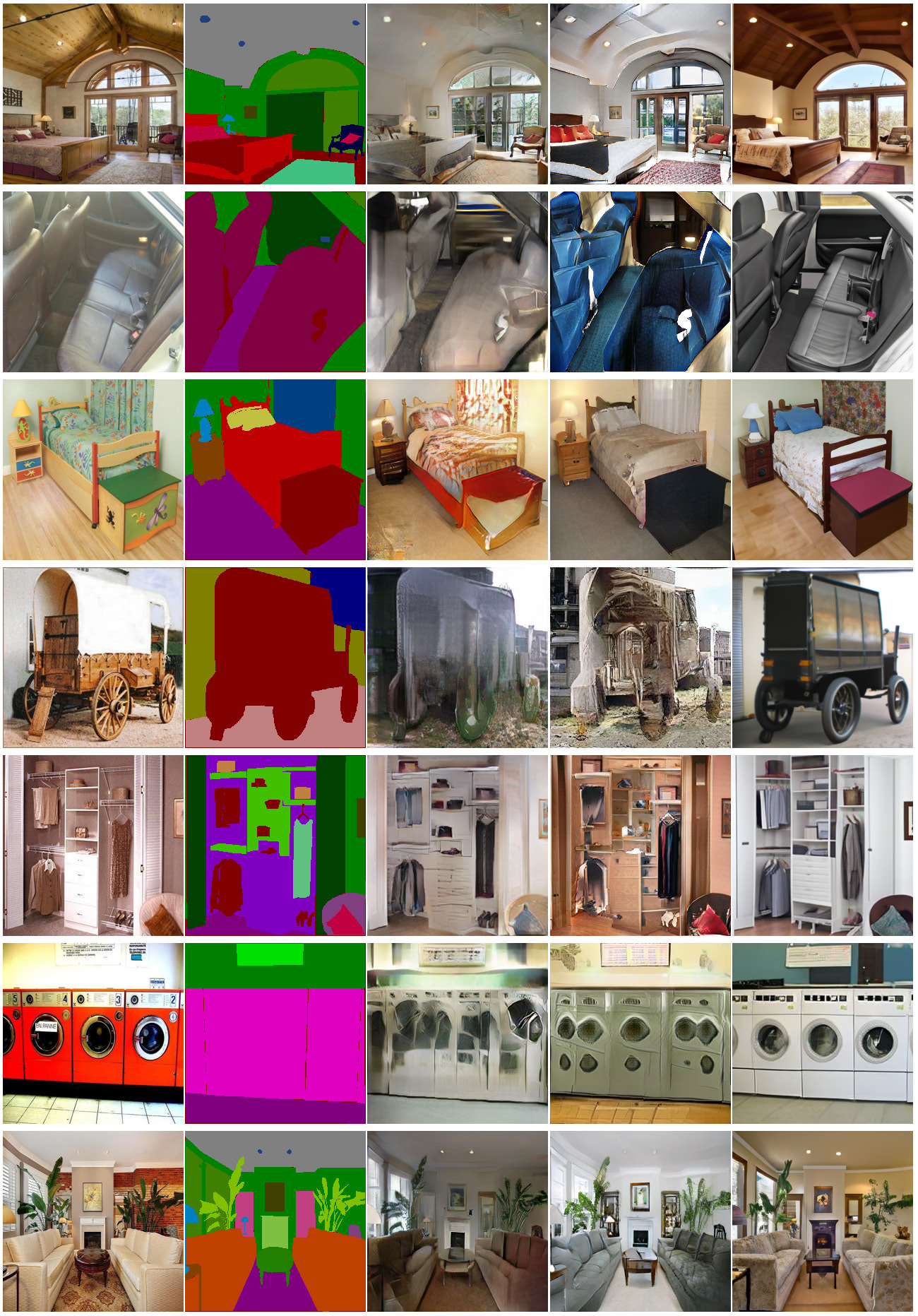} \\ 
    \begin{tabular}{@{} c@{\hspace{18mm}}c@{\hspace{18mm}}c@{\hspace{18mm}}c@{\hspace{18mm}}c@{}}
 
     GT&   Condition &SPADE &OASIS &Ours    
    \end{tabular}
   \caption{Visual comparisons on mask-to-image synthesis on ADE20K.}
    \label{fig:ade-mask1}
\end{figure*}    

\begin{figure*}[t]
    \centering 
    \small
    \includegraphics[width=\linewidth]{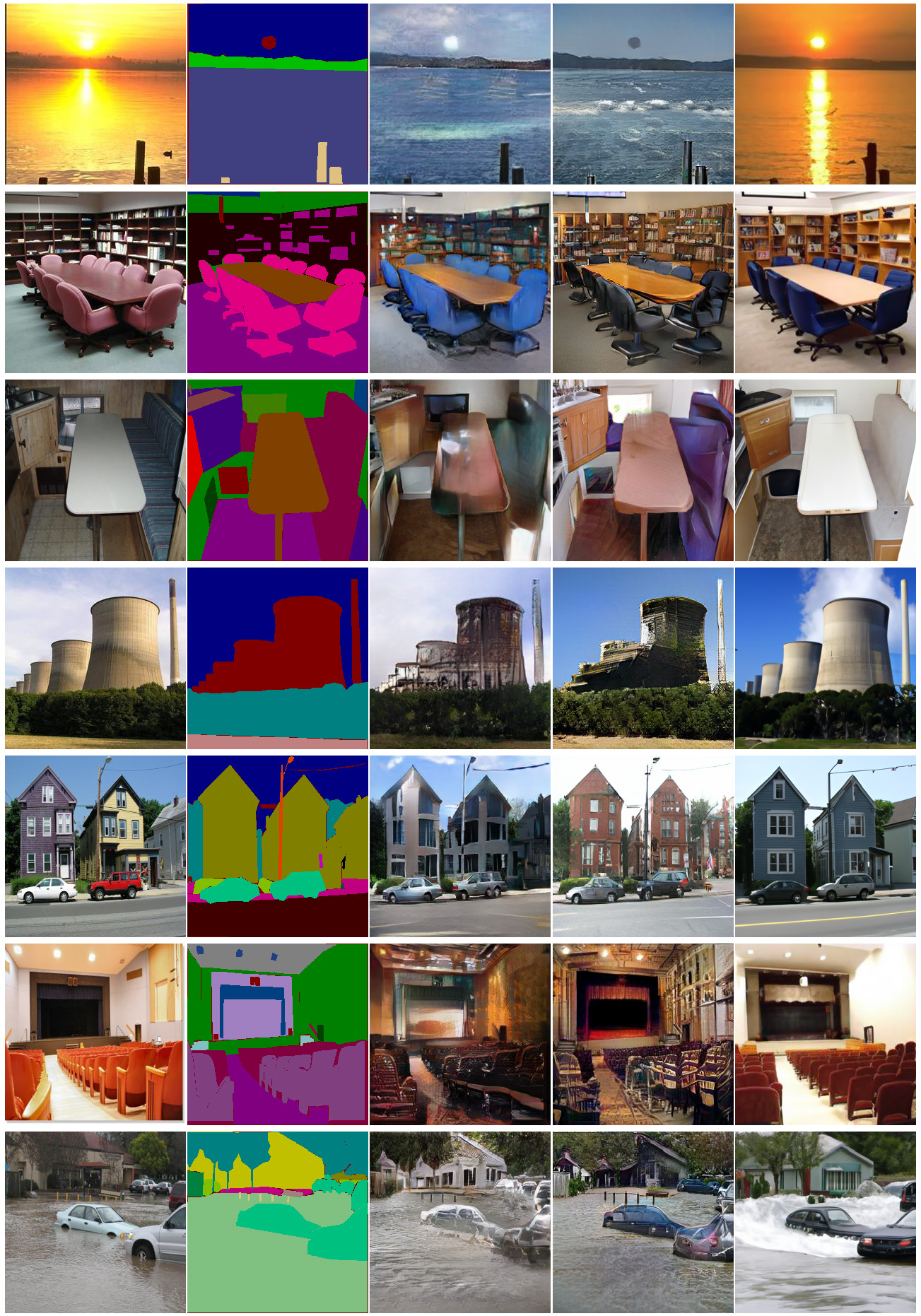} \\ 
    \begin{tabular}{@{} c@{\hspace{18mm}}c@{\hspace{18mm}}c@{\hspace{18mm}}c@{\hspace{18mm}}c@{}}

     GT&   Condition &SPADE &OASIS &Ours    
    \end{tabular}
   \caption{Visual comparisons on mask-to-image synthesis on ADE20K.}
    \label{fig:ade-mask2}
\end{figure*}   

\begin{figure*}[t]
    \centering 
    \small
    \includegraphics[width=\linewidth]{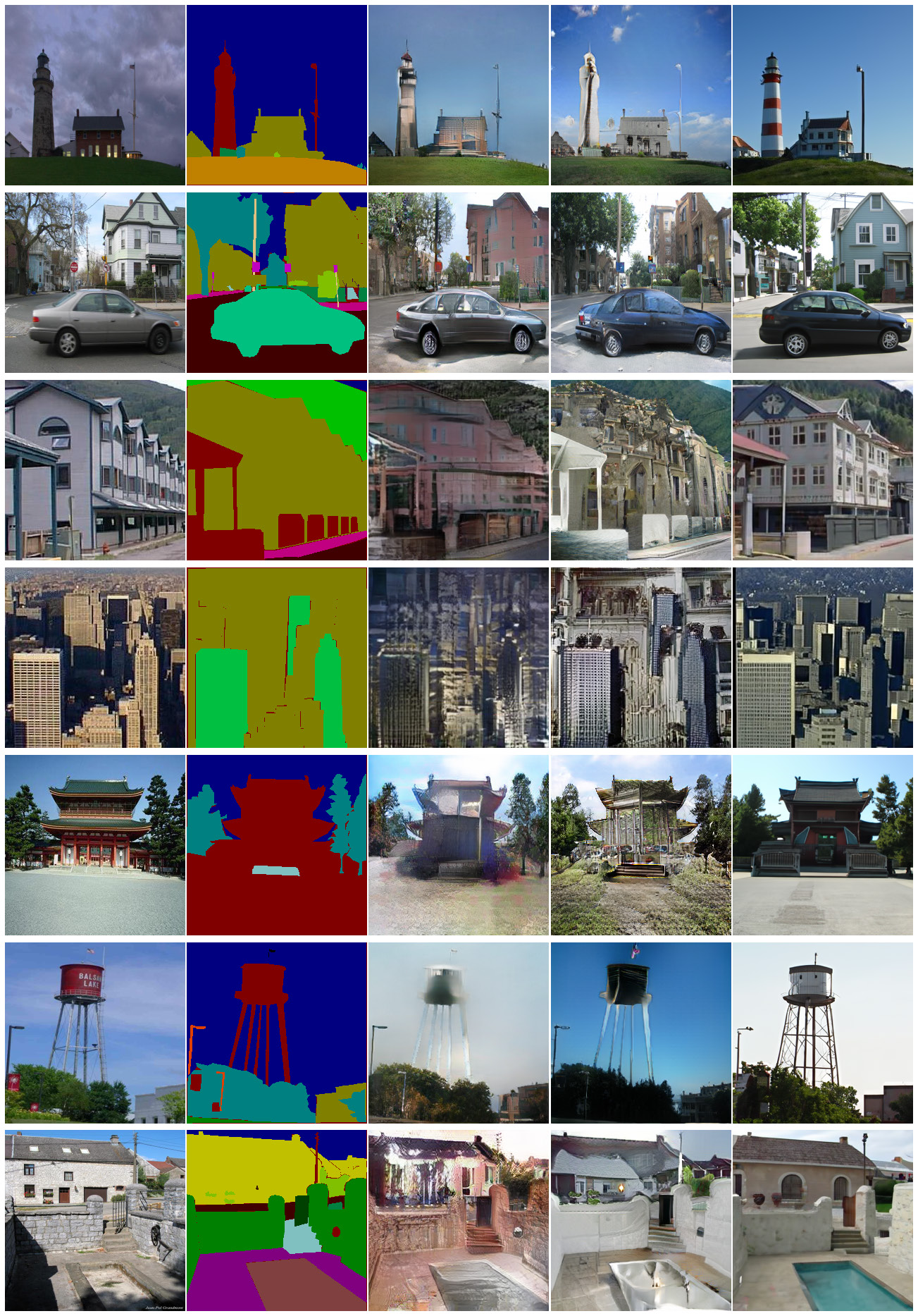} \\ 
    \begin{tabular}{@{} c@{\hspace{18mm}}c@{\hspace{18mm}}c@{\hspace{18mm}}c@{\hspace{18mm}}c@{}}
 
     GT&   Condition &SPADE &OASIS &Ours    
    \end{tabular}
   \caption{Visual comparisons on mask-to-image synthesis on ADE20K.}
    \label{fig:ade-mask3}
\end{figure*}    
 
\begin{figure*}[t]
    \centering 
    \small
    \includegraphics[width=\linewidth]{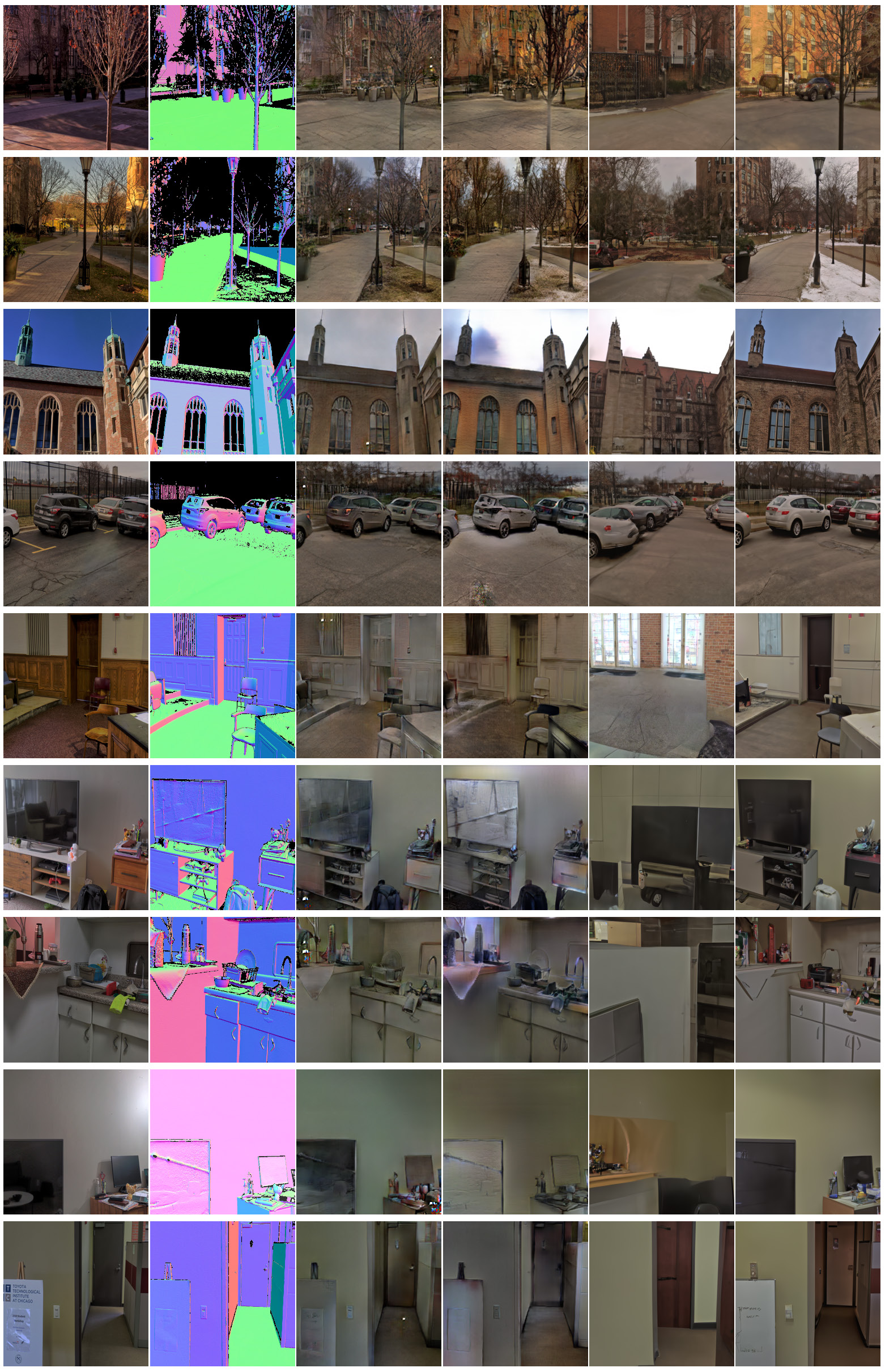} \\ 
    \begin{tabular}{@{} c@{\hspace{14mm}}c@{\hspace{10mm}}c@{\hspace{12mm}}c@{\hspace{9mm}}c@{\hspace{10mm}}c@{}}
  
     GT&   Condition & Pix2PixHD & Spade &From Scratch &Ours     
    \end{tabular}
 
   \caption{Visual comparisons on geometry-to-image synthesis on DIODE.}
    \label{fig:diode_1}
\end{figure*} 

\begin{figure*}[t]
    \centering 
    \small
    \includegraphics[width=\linewidth]{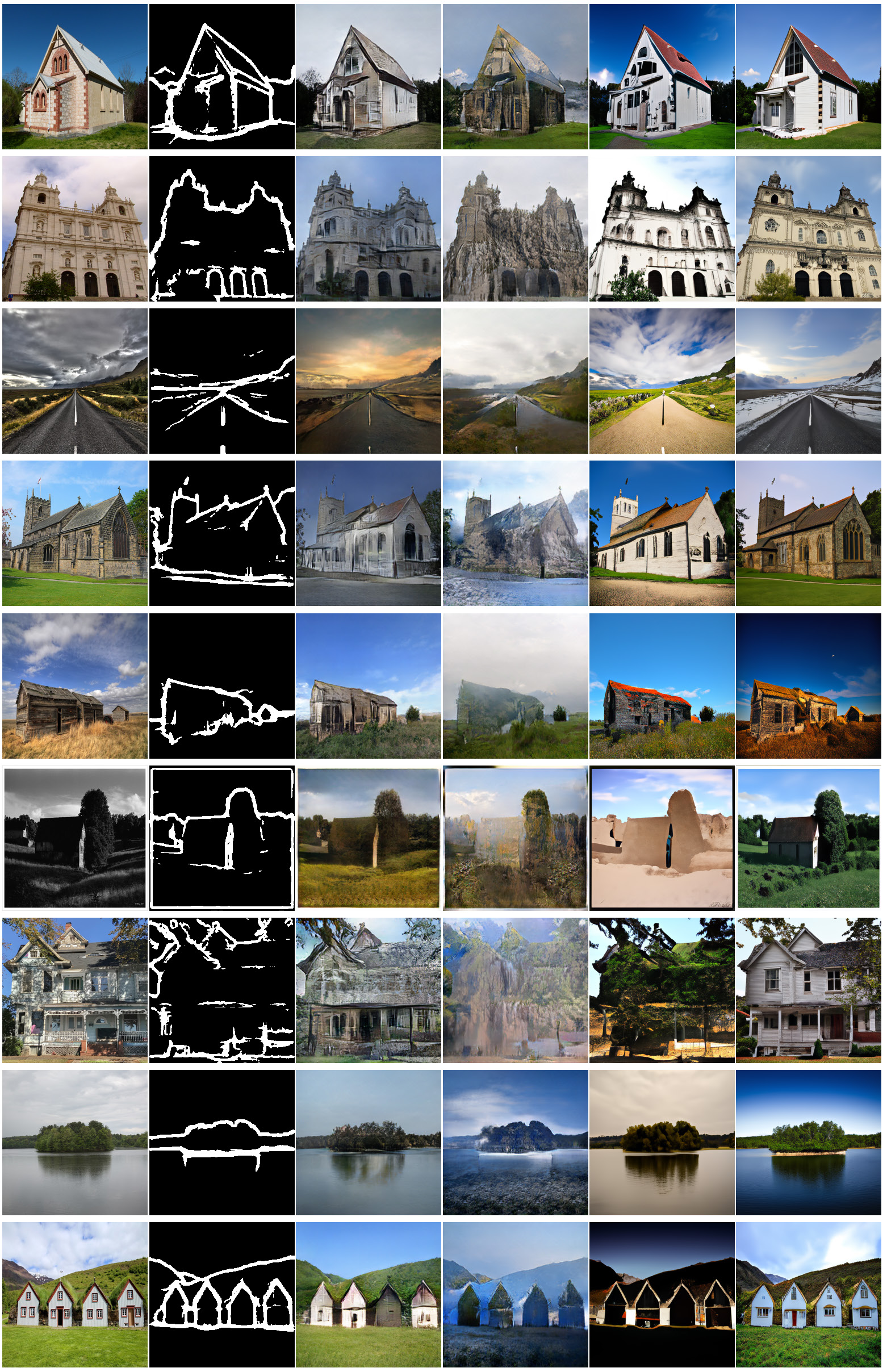} \\ 
    \begin{tabular}{@{} c@{\hspace{14mm}}c@{\hspace{10mm}}c@{\hspace{12mm}}c@{\hspace{9mm}}c@{\hspace{10mm}}c@{}}
  
     GT&   Condition & Pix2PixHD & Spade &From Scratch &Ours       
    \end{tabular}
   \caption{Visual comparisons on sketch-to-image synthesis on Flickr.}
    \label{fig:flk1}
\end{figure*} 

\end{document}